\documentclass{article}
\usepackage[preprint]{log_2026}						%

\usepackage{booktabs}						%
\usepackage{multirow}						%
\usepackage{amsfonts}						%
\usepackage{graphicx}						%
\usepackage{duckuments}						%

\usepackage[numbers,compress,sort]{natbib}	%

\usepackage{tikz}
\usepackage{pgfplots}
\pgfplotsset{compat=1.18}
\usepackage{xcolor}
\definecolor{charcoal}{RGB}{54,69,79}

\title[RAW-Explainer]{RAW-Explainer: Post-hoc Explanations of Path Reasoning Graph Neural Networks for Knowledge Graph Completion}

\author[R. Kubo et al.]{Ryoji Kubo \quad Samyam Lamichhane \quad Djellel Difallah \\
  NYU Abu Dhabi, UAE \\
  \texttt{\{ryojikubo, slamichhane, djellel\}@nyu.edu}}

\makeatletter
\renewcommand{\@noticestring}{Accepted at the Learning on Graphs Conference (LoG 2026).}
\makeatother

\begin{document}

\maketitle

\begin{abstract} Path reasoning graph neural networks have demonstrated state-of-the-art performance on knowledge graph completion tasks such as link prediction. However, interpreting their predictions remains a challenging open problem, as most GNN explainability methods target node-level or graph-level tasks. In this paper, we propose RAW-Explainer (\underline{Ra}ndom \underline{W}alk Explainer), a novel explainability method tailored to path reasoning GNNs that generates connected, concise, and interpretable subgraph explanations for link prediction in knowledge graphs. Our method uses heterogeneous information to identify factual explanations and encourages connectivity through a random walk objective, using a parameterized model for efficient collective explanation. Furthermore, RAW-Explainer overcomes the distribution shift issue when evaluating explanatory subgraphs that are orders of magnitude smaller than the full graph by proposing a robust evaluator that generalizes to the subgraph distribution. Extensive results on standard knowledge graph completion benchmarks demonstrate that our approach strikes a balance between explanation quality and computational efficiency.
Code and data are available at: {\textcolor{blue!80!black}{\url{https://github.com/colab-nyuad/RAWExplainer}}}
\end{abstract}

\section{Introduction}
\label{submission}
Knowledge graphs (KGs) are heterogeneous networks that capture real-world entities and their relationships. They play a central role in organizing, representing, and integrating structured knowledge across a variety of domains. However, KGs are often incomplete, which makes the task of KG completion (i.e., predicting missing links) essential for many knowledge graph applications. Recently, path reasoning graph neural networks (GNNs), such as NBFNet~\cite{nbfnet}, RED-GNN~\cite{redgnn}, and A*Net~\cite{zhu2024net}, have emerged as a powerful approach for KG completion, outperforming traditional rule-based and embedding-based methods. These models learn differentiable path-finding procedures, providing a strong inductive bias toward multi-hop reasoning. However, interpreting their predictions remains challenging due to complex non-linear transformations, making post-hoc explainability essential for real-world deployment.

Explaining path reasoning GNNs presents distinct challenges. To address GNN explainability, various methods have been proposed, primarily for homogeneous graph tasks such as node and graph classification~\cite{yuan2021explainability,amara2022graphframex,graphlime,DBLP:conf/icml/LinLL21,pgm-explainer}. A popular approach is the perturbation framework, in which informative subgraphs are identified by perturbing the input graph of a trained GNN in a post-hoc manner~\cite{ying2019gnnexplainer,why_should_i_trust_you}. However, many existing methods were originally developed for node-level or graph-level tasks on homogeneous graphs, and do not readily extend to edge-level tasks in heterogeneous graphs where relation types and direction must be accounted for. Moreover, in the context of knowledge graphs, connectivity of the explanatory subgraph is a natural desideratum, as it provides a coherent reasoning chain that is more human-interpretable~\cite{DBLP:conf/aaai/WangWX00C19}, yet many current methods do not guarantee it. Methods that do ensure connectivity, such as path-based approaches, incur significant computational cost and have limited expressiveness~\cite{page-link,DBLP:journals/pvldb/SunHYYW11,DBLP:journals/jasis/Liben-NowellK07}, and existing evaluations remain limited to synthetic datasets.\looseness=-1

Amara et al.~\cite{amara2022graphframex} argue that an ideal explanation should satisfy both factual and counter-factual requirements. Given that knowledge graphs are typically very large and that explanatory patterns are often confined to small paths or connected subgraphs~\cite{Bollobas2001RandomGraphs,DBLP:conf/kdd/JehW02}, we focus on uncovering factual explanations that are interpretable and support prediction of the true entity. Moreover, an effective explainer should be robust to potential out-of-distribution (OOD) issues when evaluating the informativeness of subgraphs, particularly when they are significantly smaller than the original graph~\cite{kubo2024xgexplainer,hooker2019benchmark,jethani2021learned}.

In this work, we propose RAW-Explainer (\underline{Ra}ndom \underline{W}alk Explainer), a novel explainability method for path reasoning GNNs on knowledge graph completion that addresses the aforementioned limitations. RAW-Explainer bridges subgraph-based and path-based explanation paradigms: it integrates parameterized edge-mask learning with a random walk objective, combining the flexibility of learned edge masks with the connectivity of path-based reasoning, while avoiding the computational overhead typically associated with the latter. RAW-Explainer naturally incorporates heterogeneous information such as relation types and direction, and introduces a robust evaluator that overcomes OOD issues to assess the informativeness of subgraphs during the perturbation process. We employ a realistic end-to-end evaluation framework and compare against state-of-the-art explainers on four large knowledge graph datasets. The results show that RAW-Explainer yields better explanatory subgraphs under limited edge budgets and is significantly more efficient than existing baselines.\looseness=-1

Our key contributions are summarized as follows:
\begin{itemize}
    \item \textit{A unified subgraph-based and path-based GNN explanation method:}
    By integrating parameterized edge-mask learning with random walks, RAW-Explainer encourages connectivity in local subgraph explanations while remaining computationally efficient.\looseness=-1

    \item \textit{A robust evaluation framework for perturbation-based explainers:}
    We propose a novel distance-based model evaluator that approximates the underlying graph distribution, thereby mitigating OOD evaluation biases and yielding more faithful explanation quality estimates.\looseness=-1

    \item \textit{Thorough empirical evaluation:}
    Using a factual evaluation protocol on four knowledge graph benchmarks, RAW-Explainer consistently outperforms state-of-the-art path-based explainability baselines in efficiency, faithfulness, and structural coherence of generated explanations.\looseness=-1
\end{itemize}
\section{Related Work} \label{related}
GNNs have emerged as a powerful class of machine learning algorithms for relational graphs, iteratively transforming node embeddings through message passing and neighborhood aggregation. Path reasoning models, including NBFNet~\cite{nbfnet}, RED-GNN~\cite{redgnn}, ULTRA~\cite{galkin2023ultra}, and A*Net~\cite{zhu2024net}, leverage query-dependent transformations and have demonstrated superior performance in knowledge graph completion. Understanding how these models reason is essential for real-world applications and often requires post-hoc explainability. Alternative approaches such as NeuralLP~\cite{yang2017neurallp} and DRUM~\cite{sadeghian2019drum} learn logical rules directly over knowledge graphs, offering built-in interpretability, but generally underperform compared to state-of-the-art GNNs.
\begin{table}[t]
\centering
\footnotesize %
\caption{
    GNN explanation methods for KG completion.
    Het.: heterogeneous;
    Conn.: connectivity;
    Path: path-aware;
    OOD: OOD robust;
    Ind.: inductive;
    Scal.: scalable;
    LP: link prediction.
    $\checkmark$: supported;
    ($\checkmark$): non-trivial extension;
    $\times$: not supported.
}
\label{tab:status}
\begin{tabular*}{\columnwidth}{@{\extracolsep{\fill}}lccccccc}
\toprule
\rule{0pt}{2.2em}

Method & Het. & Conn. & Path & OOD & Ind. & Scal. & LP \\

\midrule
GNN-Expl.~\cite{ying2019gnnexplainer} {\tiny NeurIPS'19} & $\times$ & $\times$ & $\times$ & $\times$ & $\times$ & $\times$ & ($\checkmark$) \\
PG-Expl.~\cite{PGExplainer} {\tiny NeurIPS'20} & $\times$ & $\times$ & $\times$ & $\times$ & $\checkmark$ & $\checkmark$ & ($\checkmark$) \\
PaGE-Link~\cite{page-link} {\tiny WWW'23} & $\checkmark$ & $\checkmark$ & $\checkmark$ & $\times$ & $\times$ & $\times$ & $\checkmark$ \\
PowerLink~\cite{powerlink} {\tiny KDD'24} & $\checkmark$ & $\checkmark$ & $\checkmark$ & $\times$ & $\times$ & ($\checkmark$) & $\checkmark$ \\
\midrule
\textbf{RAW-Expl.} & $\checkmark$ & $\checkmark$ & $\checkmark$ & $\checkmark$ & $\checkmark$ & $\checkmark$ & $\checkmark$ \\
\bottomrule
\end{tabular*}
\end{table}

Several GNN explainability methods have been developed to address the challenge of interpreting GNN predictions~\cite{yuan2021explainability,agarwal2023evaluating,kakkad2023surveyexplainabilitygraphneural,GstarX}. GNN-Explainer~\cite{ying2019gnnexplainer} is a pioneering method that explains each instance by optimizing a mask over the adjacency matrix using mutual information. PG-Explainer~\cite{PGExplainer} learns a parameterized model to generate explanations across multiple instances using a factual objective, enabling inductive generalization at inference. XG-Explainer~\cite{kubo2024xgexplainer} focuses on alleviating the OOD issue of subgraphs during the perturbation process by training an alternative GNN evaluator on randomly sampled subgraphs, while MixupExplainer~\cite{mixupexplainer} does so via data augmentation.

In the field of knowledge graph completion explanation, only a handful of methods exist. \autoref{tab:status} summarizes key approaches relevant to our setup. PaGE-Link~\cite{page-link} is tailored for heterogeneous link prediction, learning a mask over the adjacency matrix for each instance and finding informative paths as explanations with a factual objective and path-enforcing mask learning. Power-Link~\cite{powerlink} employs a graph-powering technique to directly extract informative paths. While originally proposed for node-level and graph-level tasks, GNN-Explainer and PG-Explainer can be reformulated for link prediction by optimizing masks with respect to edge-level queries.

Another line of work targets embedding-based link prediction models (e.g., TransE, DistMult) and is not applicable to GNN-based predictors: KGEx~\cite{KGEx} and XKE~\cite{RUSCHEL2024109183} attribute predictions to individual embedding dimensions, KGExplainer~\cite{ma2025towards} uses a distilled proxy evaluator to score subgraph explanations, and Kelpie~\cite{DBLP:journals/pvldb/RossiFMT22} identifies influential sets of training facts that support or refute a prediction.\looseness=-1

RAW-Explainer extends the benefits of robust evaluation and parameterized global explanation to heterogeneous graphs while drawing inspiration from existing methods to generate local, interpretable subgraphs in a scalable manner.

\section{Preliminaries}
\label{prelim}

A knowledge graph (KG) is defined as $\mathcal{G} = (\mathcal{V}, \mathcal{E}, \mathcal{R}),$
where $\mathcal{V}$ is the set of entities, $\mathcal{R}$ is the set of relations, and $\mathcal{E}$ is the set of edges (or triples). Each edge corresponds to a triple $(h, r, t)$, with $h, t \in \mathcal{V}$ and $r \in \mathcal{R}$. The task of KG completion is to identify the missing tail or head entity for a query, i.e., either $(h, r, ?)$ or $(?, r, t)$, such that the triple $(h, r, t)$ holds true in $\mathcal{G}$. By incorporating inverse relations, all queries can be uniformly represented in the form $(h, r, ?)$; for simplicity, we denote a query as $(h, r)$.

A GNN model $\Phi$ learns the conditional distribution $P_\Phi(\mathbf{Y}|\mathcal{G}, (h, r))$, where $\mathbf{Y}$ is a binary random variable over the entities $v \in \mathcal{V}$, by iteratively transforming node embeddings through message passing and aggregation across $L$ layers, and scoring candidate tails through a decoder.

When obtaining a \emph{factual explanation} for a query $(h, r)$, the task is to find the most informative small subgraph $\mathcal{G}_S \subseteq \mathcal{G}$ that retains the true label, formally $\operatorname*{arg\,max}_{\mathcal{G}_S} P_\Phi(\mathbf{Y}|\mathcal{G}_S, (h, r))$ subject to $|\mathcal{G}_S| \leq k$, where $|\cdot|$ denotes the number of edges and $k$ is the user-defined budget. We adopt a \textit{phenomenon}-focused~\cite{amara2022graphframex} explanation context by setting our objective to retain the true label $\mathbf{Y}$ rather than the original prediction $\mathbf{\hat{Y}}$ made by the GNN given $\mathcal{G}$. Model-focused and counter-factual objectives are discussed in Appendix~\ref{app:other-objectives}.
\section{Proposed Method: RAW-Explainer}
\label{method}

We design RAW-Explainer to efficiently produce small, explanatory subgraphs for knowledge graph completion. Our approach is based on the following design principles: (1) Encourage connectivity in the output explanations for better interpretability.
(2) Enable global explanation capabilities at test time by employing a parameterized model. (3) Accurately evaluate the informativeness of subgraphs during the perturbation process using a robust GNN evaluator.
RAW-Explainer achieves these goals through three key components: a heterogeneous mask learner, a random-walk-based connectivity constraint, and a robust GNN evaluator.

\subsection{Heterogeneous Mask Learner}
We introduce a heterogeneous mask learner that determines the importance of both entities and relations in a knowledge graph. Building on global explanation techniques such as PG-Explainer~\cite{PGExplainer}, we employ a parameterized model capable of learning edge importance across multiple instances.

\subsubsection{Parameterized Soft-Mask Model} Given a knowledge graph $\mathcal{G}$ and a query $(h, r)$, a pre-trained GNN model $\Phi$ is used to compute $d$-dimensional embeddings $\mathbf{Z}_E \in \mathbb{R}^{|\mathcal{V}| \times d}$ and $\mathbf{Z}_R \in \mathbb{R}^{|\mathcal{R}| \times d}$ for entities and relations, respectively.
For each edge $(s, p, o) \in \mathcal{E}$, we determine its importance with respect to the query $(h, r)$ by concatenating the embeddings of the source entity $s$, the relation $p$, the target entity $o$, the query head $h$, and the query relation $r$. This concatenated vector is then passed through an MLP $\eta$:
\begin{equation}
\label{edge_imp}
\omega_{s, p, o} = \eta([\mathbf{z}_{s}; \mathbf{z}_{p}; \mathbf{z}_{o}; \mathbf{z}_{h}; \mathbf{z}_{r}]).
\end{equation}
Subsequently, the Gumbel-Softmax function is applied to $\omega_{s, p, o}$ to constrain its value to the interval $[0, 1]$, providing a differentiable approximation of discrete edge sampling. The collection of all edge scores forms the soft mask $\mathbf{\Omega}$ of the graph.

\subsubsection{Soft-Mask Evaluator} Next, to assess the relevance of the learned mask, we use an evaluation module, a GNN $\Phi_{eval}$, that performs link prediction using the masked graph:
\begin{equation}
\label{masked_pred}
\hat{\mathbf{Y}}_{\mathbf{\Omega}} = \Phi_{eval}(\mathcal{G}, (h, r), \mathbf{\Omega}).
\end{equation}
This step assesses how well the mask $\mathbf{\Omega}$ captures the important edges necessary for answering the query. Although the backbone GNN $\Phi$ could be used as the evaluator $\Phi_{eval}$, we explore the limitations of this choice and present an alternative in \autoref{sec:evalrob}.

Rather than using the standard GNN update, we integrate a dynamic masking mechanism into each layer, where the message passed from each neighboring node is scaled by its corresponding mask value:
\begin{equation}
\label{masked_iter}
\mathbf{h}_s^l = \zeta\left(\mathbf{h}_s^{l-1}, \, \bigoplus\limits_{o \in N_s}^{\mathbf{\Omega}} \omega_{s, p, o} \cdot \psi\big(\mathbf{h}_s^{l-1}, \mathbf{h}_o^{l-1}, p\big)\right).
\end{equation}
Here, $\psi$ computes the message from neighbor $o$ to node $s$, and $\zeta$ updates the node representation. The operator $\bigoplus^{\mathbf{\Omega}}$ denotes a dynamic aggregation that incorporates the mask values. For example, if mean aggregation is used, the normalization term adjusts based on the effective number of neighbors with significant mask values.

The loss term for the heterogeneous mask learner is computed as the binary cross-entropy between the label $\mathbf{Y}$ and the prediction $\hat{\mathbf{Y}}_{\mathbf{\Omega}}$ obtained using the mask $\mathbf{\Omega}$:
\begin{equation}
\label{pred_loss}
    \mathcal{L}_{pred}(\mathbf{\Omega}) = BCE({\mathbf{Y}}, \hat{\mathbf{Y}}_{\mathbf{\Omega}})
\end{equation}

\subsection{Robust GNN Evaluator} 
\label{sec:evalrob}
Recent studies have shown that perturbation-based frameworks can be undermined by an out-of-distribution (OOD) issue~\cite{kubo2024xgexplainer,jethani2021learned, hooker2019benchmark, mixupexplainer}. In the case of GNNs, this occurs because the explanatory subgraph $\mathcal{G}_S$ can fall outside the original training graph distribution, so the backbone GNN predictor $\Phi$ may not make accurate predictions to effectively train the learnable mask. Existing solutions create an auxiliary GNN evaluator model $\Phi_{eval}$ trained with uniform edge-drop probability on the training graph, simulating OOD conditions so the evaluator generalizes better to perturbed subgraphs. Although this has primarily been studied in node and graph classification, the findings strongly suggest that a robust GNN evaluator is just as critical for link prediction, which we validate in \autoref{sec:gnn_eval_exp}.

However, directly applying uniform edge-drop augmentation to knowledge graphs has inherent limitations: explanatory subgraphs for knowledge graph queries are usually localized and highly relevant to specific entities or relations, so uniform perturbations may introduce irrelevant or overly broad OOD contexts. To overcome this, we propose a distance-based robust evaluator $\Phi_{eval}^{dist}$, trained by assigning edge-dropping probabilities that increase with the edge distance from query entities up to a maximum, thus preserving the contextual relevance and locality of explanatory subgraphs, while $\Phi$ remains the target model. We evaluate its effectiveness in \autoref{sec:gnn_eval_dist_exp}.

\subsection{Structural Constraint}

We use a structural objective to encourage connectivity in the subgraph selected by $\mathbf{\Omega}$, as connectivity captures paths from the head to candidate tails via key intermediary nodes. To achieve this, we implement a random-walk-assisted approach that rewards edges within the subgraph induced by highly ranked nodes and penalizes edges outside it.

To perform the random walk efficiently, we use power iterations on a sparse matrix representation of $\mathcal{G}$. We aggregate the relational dimension of $\mathbf{\Omega}$ (using summation by default) to obtain $\mathbf{\Omega}' \in \mathbb{R}^{|\mathcal{V}| \times |\mathcal{V}|}$, where $\mathbf{\Omega}'_{ij}$ contains the aggregated weight of edges from source $j$ to destination $i$; full relational information is retained in the mask learner and evaluator. Only existing edges are stored, with self-loops added for isolated entities. We use \autoref{stoch_adj} to form a column-stochastic matrix $\mathbf{\Omega}_S$, so the transition probabilities from each source sum to 1.
\begin{equation}
\label{stoch_adj}
    \mathbf{\Omega}_S = (1 - \alpha) \cdot \sigma_{\mathrm{sp}}(\mathbf{\Omega}') + \alpha \cdot \sigma_{\mathrm{sp}}(\boldsymbol{\tau})
\end{equation}
where $\alpha$ is the teleportation probability and $\sigma_{\mathrm{sp}}$ applies softmax only over the stored entries of each column; absent entries receive zero probability. We form a seed list from the query head $h$ and the top $m$ tails predicted by $\Phi_{eval}$ on the full graph $\mathcal{G}$. Let $\rho$ be the set of entities in this list and $c_i$ the number of occurrences of entity $i$. The sparse teleportation matrix $\boldsymbol{\tau} \in \mathbb{R}^{|\mathcal{V}| \times |\mathcal{V}|}$ stores $\boldsymbol{\tau}_{ij}=c_i$ for $i\in\rho$ and every source $j$, with all other entries absent. Thus teleportation is restricted to $\rho$ and is uniform over these entities when the seeds are distinct.

Given $\mathbf{\Omega}_S$, we can initiate the power iteration process for the random walk. Specifically, we use the personalized PageRank (PPR) algorithm~\cite{pagerank} such that the obtained node distribution is query-dependent. The initial node distribution $\boldsymbol{\pi}^0 \in \mathbb{R}^{|\mathcal{V}|}$ is defined as $\boldsymbol{\pi}_i^0 = c_i/(m+1)$, assigning mass only to the seed entities. We repeat the iteration \mbox{$\boldsymbol{\pi}^{t} = \mathbf{\Omega}_S \cdot \boldsymbol{\pi}^{t-1}$} until the node distribution converges, i.e. \mbox{$\|\boldsymbol{\pi}^{t} - \boldsymbol{\pi}^{t-1}\|_1 \leq \epsilon$}.

After convergence, we retrieve the top $l$ entities from the final node distribution $\boldsymbol{\pi}^{T}$ and define $\mathcal{E}_{in} = \{(s,p,o)\in\mathcal{E} : \{s,o\}\subseteq\mathrm{top}\text{-}l(\boldsymbol{\pi}^{T})\}$ as edges within the walked subgraph and $\mathcal{E}_{out} = \mathcal{E} \setminus \mathcal{E}_{in}$ as its complement. The loss is then computed as:
\begin{equation}
\label{rw_loss}
    \mathcal{L}_{PPR}(\mathbf{\Omega}) = - \frac{\beta_{in}}{|\mathcal{E}_{in}|}\sum\limits_{e\in\mathcal{E}_{in}}\ \omega_e + \frac{\beta_{out}}{|\mathcal{E}_{out}|}\sum\limits_{e\in\mathcal{E}_{out}}\ \omega_e
\end{equation}
where $\beta_{in}, \beta_{out}$ weight the mean edge importance inside and outside the walked subgraph, respectively. During training, each mean is computed over the corresponding edges pooled across the batch. Using this loss in conjunction with the prediction loss in \autoref{pred_loss} encourages the explainer to retrieve a connected subgraph explored by PPR as the explanation.

At inference, RAW-Explainer selects the top entities/edges from $\boldsymbol{\pi}^{T}$ to sample the explanation $\mathcal{G}_S$ that satisfies the user-defined budget, $|\mathcal{G}_S|~\leq~k$.

\subsection{Optimization}
Since we seek to generate concise explanatory subgraphs, we introduce regularization terms $\mathcal{R}(\mathbf{\Omega})$ that penalize the size of $\mathbf{\Omega}$ and encourage an entropy-based approximation of discrete sampling. These regularizers push the explainer toward learning a sparse mask, akin to a binary mask rather than a dense or continuous one, following similar techniques in prior work~\cite{ying2019gnnexplainer,PGExplainer,page-link}. In practice, the size penalty ensures that only a few edges receive high importance scores, and the entropy penalty nudges mask values to be near 0 or 1, making them easier to interpret.

While we optimize $\eta$, the parameters of the evaluator $\Phi_{eval}$ remain fixed. If we were to optimize $\Phi_{eval}$ jointly with $\eta$, the model could adapt its internal representations to rely on edges even when their mask values are low~\cite{jethani2021learned}. By freezing $\Phi_{eval}$, we ensure that edge importance truly reflects the features of the current model rather than newly learned dependencies, thus preserving the semantics of the explanation.

RAW-Explainer is optimized by combining the prediction loss from \autoref{pred_loss}, the structural constraint from \autoref{rw_loss}, and the regularization term: 
\begin{equation}
\mathcal{L}_{rawexp}(\mathbf{\Omega}) = \mathcal{L}_{pred}(\mathbf{\Omega}) + \mathcal{L}_{PPR}(\mathbf{\Omega}) + \mathcal{R}(\mathbf{\Omega}).
\end{equation}

\begin{figure*}[t]
  \centering
  \resizebox{0.9\textwidth}{!}{%
    \includegraphics{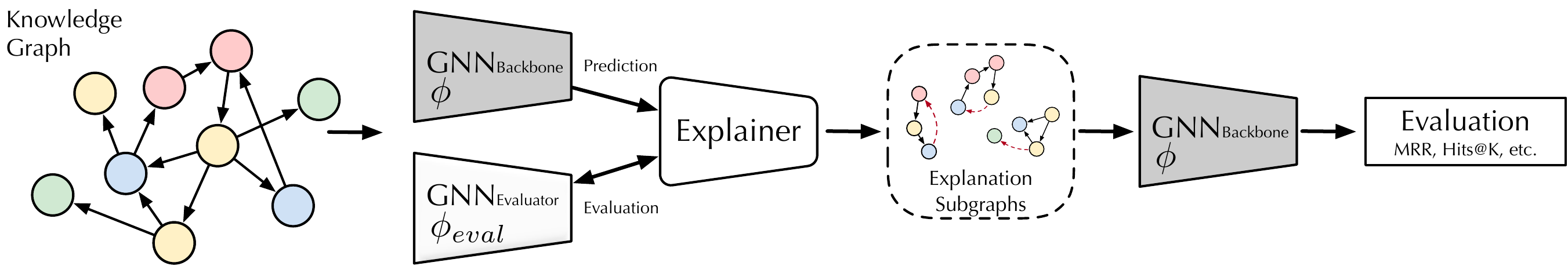}
  }
  \caption{End-to-end factual evaluation protocol for GNN explanations in KG completion.}
  \label{fig:sys}
\end{figure*}

\subsection{Protocol for Evaluating GNN Explanations}
\label{sec:framework}
There is currently a lack of consensus on how to evaluate knowledge graph explanations, particularly when the ground-truth explanatory subgraphs are unavailable. To address this gap, we propose a phenomenon-focused factual explanation evaluation framework that assesses whether the generated explanations retain enough information to predict the true entity. \autoref{fig:sys} illustrates our end-to-end evaluation protocol: an explainer is trained to generate explanatory subgraphs of a fixed edge budget, which are then fed back to the backbone GNN and evaluated using standard knowledge graph completion metrics such as mean reciprocal rank (MRR) and Hits@K. Since all explainers are evaluated under identical conditions, the framework provides a fair comparative basis despite the inherent OOD challenges, and we further vary the edge budget to assess the stability of rankings across different explanation sizes.
\section{Experiments}
\label{experiments}
We evaluate RAW-Explainer through a series of experiments to answer the following research questions. \textbf{RQ1:} Does distribution shift undermine a GNN model as a perturbation-based evaluator? \textbf{RQ2:} Can a distance-aware robust evaluator better handle the locality of KG explanations? \textbf{RQ3:} Does RAW-Explainer produce more faithful and efficient explanations compared to state-of-the-art baselines?

\subsection{Experimental Setup} 

\subsubsection{Datasets} We use four real-world knowledge graph completion datasets: FB15k-237~\cite{toutanova-chen-2015-observed}, WN18RR~\cite{Dettmers_Minervini_Stenetorp_Riedel_2018}, NELL-995~\cite{DBLP:conf/emnlp/XiongHW17}, and YAGO3-10~\cite{DBLP:conf/www/SuchanekKW07}. These datasets vary considerably in scale, density, and number of relation types, providing a diverse testbed for evaluating explainability methods (see \autoref{tab:dataset-stats} for statistics). To facilitate bidirectional traversal, we augment all datasets with inverse relations.

\subsubsection{GNN Backbone} We adopt \textbf{NBFNet}~\cite{nbfnet}, a state-of-the-art GNN architecture for knowledge graph completion. Specifically, NBFNet operates by learning a differentiable method for pathfinding within the graph, providing a clear inductive bias toward path-based reasoning. This makes it an ideal testbed for studying explainability, as its path-based nature offers a stronger intuition about the structural patterns it captures, enabling more grounded validation of the generated explanations. While RAW-Explainer is model-agnostic by design, we focus on NBFNet to thoroughly investigate explanation quality; an additional evaluation using RED-GNN is reported in \autoref{app:backbone-sensitivity}. We follow the hyperparameter configurations detailed in the original paper. Unless otherwise stated, the term \textit{backbone GNN} or $\Phi$ refers to the NBFNet model trained on our datasets.

\subsubsection{Evaluation} For the task of link prediction, where a tail-prediction query $(h, r, ?)$ or head-prediction query $(?, r, t)$ ranks all nodes in $\mathcal{V}$, we follow the standard filtered ranking protocol~\cite{transe} and rank the true triple $(h, r, t)$ against all negative triples $(h, r, t')$ or $(h', r, t)$ that do not exist in the KG. We report the MRR as a concise evaluation metric, averaged over both head and tail prediction tasks.

\subsection{Impact of Distribution Shift on GNN Evaluation (RQ1)}
\label{sec:gnn_eval_exp}
In this experiment, we compare two GNN model evaluators: the backbone model $\Phi$, trained exclusively on the full training graph, and an evaluation model $\Phi_{eval}$, trained on the same graph but with uniformly sampled edges removed. The goal is to examine whether $\Phi_{eval}$ provides more robust predictions under distribution shift due to smaller subgraphs.

\subsubsection{Experimental Setup} Ideally, in scenarios with known ground-truth edge importance, a robust evaluation model should sustain its performance when unimportant edges are removed from the graph $\mathcal{G}$. However, since ground-truth edge importance is unavailable for real-world datasets, we adopt a proxy strategy: we randomly remove edges from the full graph $\mathcal{G}$ at increasing drop rates and observe each model's performance degradation. Robustness is assessed based on the evaluator's ability to sustain its performance on these modified graphs. We conduct these evaluations on the validation set.

\subsubsection{Results}
\autoref{gnn_eval_result} and \autoref{tab:mrr-vs-random-drop} illustrate the results. While both $\Phi$ and $\Phi_{eval}$ show comparable performance on the full graph, their performance degrades as the probability of random edge removal increases. This decrease in performance aligns with expectations, as crucial informative edges for accurate predictions may be inadvertently removed during random edge-dropping. However, the evaluation model $\Phi_{eval}$ consistently maintains higher accuracy after edge removal at varying levels of sparsity compared to the backbone model $\Phi$. Given that only a small subset of edges in $\mathcal{G}$ significantly contributes to answering queries \cite{page-link,zhu2024net}, random edge removal predominantly affects non-informative edges, making this a meaningful proxy for the distribution shift encountered during explanation evaluation. The capacity of $\Phi_{eval}$ to maintain higher performance under these conditions is critical for accurately evaluating both factual and counter-factual aspects of explanations, supporting our claim that $\Phi_{eval}$ is a more reliable evaluator than $\Phi$.

\begin{figure}[th!]
\centering
\ref{sharedlegend0}\\[1pt]
\noindent
\begin{minipage}[t]{0.24\columnwidth}
    \centering
    \begin{tikzpicture}
      \begin{axis}[
        name=plot,
        width=1.2\linewidth,
        height=0.85\linewidth,
        xlabel={Random edge-drop probability},
        ylabel={\scriptsize MRR},
        ylabel style={yshift=-2pt},
        xmin=0, xmax=0.5,
        ymin=0.15, ymax=0.6,
        xtick={0,0.1,0.2,0.3,0.4,0.5},
        ytick={0.2,0.3,0.4,0.5,0.6},
        tick label style={font=\scriptsize},
        xlabel style={font=\scriptsize,yshift=1ex},
        legend to name=sharedlegend0,
        legend columns=2,
        legend style={
          draw=none,
          font=\scriptsize,
          column sep=0.1cm,
        },
        ymajorgrids=true,
        grid style=dashed,
      ]
        \addplot[ color=blue, mark=o, mark size=1.5pt ] coordinates {
          (0,0.551)(0.05,0.488)(0.1,0.437)
          (0.2,0.353)(0.3,0.288)(0.4,0.236)(0.5,0.196)
        };
        \addplot[ color=red, mark=square, mark size=1.5pt ] coordinates {
          (0,0.553)(0.05,0.524)(0.1,0.496)
          (0.2,0.444)(0.3,0.389)(0.4,0.341)(0.5,0.289)
        };
        \legend{$\Phi$, $\Phi_{eval}$}
      \end{axis}
      \node[below=1pt, font=\scriptsize\bfseries] at (plot.below south) {(a) WN18RR.};
    \end{tikzpicture}
  \end{minipage}%
  \hfill
  \begin{minipage}[t]{0.24\columnwidth}
    \centering
    \begin{tikzpicture}
      \begin{axis}[
        name=plot,
        width=1.2\linewidth,
        height=0.85\linewidth,
        xlabel={Random edge-drop probability},
        xmin=0, xmax=0.5,
        ymin=0.3, ymax=0.45,
        xtick={0,0.1,0.2,0.3,0.4,0.5},
        ytick={0.30,0.35,0.40,0.45},
        tick label style={font=\scriptsize},
        xlabel style={font=\scriptsize,yshift=1ex},
        ymajorgrids=true,
        grid style=dashed,
      ]
        \addplot[ color=blue, mark=o, mark size=1.5pt ] coordinates {
          (0,0.422)(0.05,0.411)(0.1,0.402)
          (0.2,0.383)(0.3,0.361)(0.4,0.337)(0.5,0.310)
        };
        \addplot[ color=red, mark=square, mark size=1.5pt ] coordinates {
          (0,0.427)(0.05,0.419)(0.1,0.412)
          (0.2,0.396)(0.3,0.382)(0.4,0.363)(0.5,0.342)
        };
      \end{axis}
      \node[below=1pt, font=\scriptsize\bfseries] at (plot.below south) {(b) FB15k-237.};
    \end{tikzpicture}
  \end{minipage}%
  \hfill
  \begin{minipage}[t]{0.24\columnwidth}
    \centering
    \begin{tikzpicture}
      \begin{axis}[
        name=plot,
        width=1.2\linewidth,
        height=0.85\linewidth,
        xlabel={Random edge-drop probability},
        ylabel style={yshift=-2pt},
        xmin=0, xmax=0.5,
        ymin=0.0, ymax=0.5,
        xtick={0,0.1,0.2,0.3,0.4,0.5},
        ytick={0.0,0.1,0.2,0.3,0.4,0.5},
        tick label style={font=\scriptsize},
        xlabel style={font=\scriptsize,yshift=1ex},
        ymajorgrids=true,
        grid style=dashed,
      ]
        \addplot[ color=blue, mark=o, mark size=1.5pt ] coordinates {
          (0,0.462)(0.1,0.130)
          (0.2,0.099)(0.3,0.087)(0.4,0.072)(0.5,0.057)
        };
        \addplot[ color=red, mark=square, mark size=1.5pt ] coordinates {
          (0,0.472)(0.1,0.455)
          (0.2,0.436)(0.3,0.414)(0.4,0.390)(0.5,0.364)
        };
      \end{axis}
      \node[below=1pt, font=\scriptsize\bfseries] at (plot.below south) {(c) NELL-995.};
    \end{tikzpicture}
  \end{minipage}%
  \hfill
  \begin{minipage}[t]{0.24\columnwidth}
    \centering
    \begin{tikzpicture}
      \begin{axis}[
        name=plot,
        width=1.2\linewidth,
        height=0.85\linewidth,
        xlabel={Random edge-drop probability},
        xmin=0, xmax=0.5,
        ymin=0.25, ymax=0.60,
        xtick={0,0.1,0.2,0.3,0.4,0.5},
        ytick={0.30,0.40,0.50,0.60},
        tick label style={font=\scriptsize},
        xlabel style={font=\scriptsize,yshift=1ex},
        ymajorgrids=true,
        grid style=dashed,
      ]
        \addplot[ color=blue, mark=o, mark size=1.5pt ] coordinates {
          (0,0.578)(0.1,0.514)(0.2,0.464)
          (0.3,0.407)(0.4,0.346)(0.5,0.283)
        };
        \addplot[ color=red, mark=square, mark size=1.5pt ] coordinates {
          (0,0.572)(0.1,0.532)(0.2,0.489)
          (0.3,0.437)(0.4,0.397)(0.5,0.349)
        };
      \end{axis}
      \node[below=1pt, font=\scriptsize\bfseries] at (plot.below south) {(d) YAGO3-10.};
    \end{tikzpicture}
  \end{minipage}%
  
\caption{Performance evaluation of $\Phi$ and $\Phi_{eval}$ under different probabilities of random edge removal.}
\label{gnn_eval_result}
\end{figure}

\subsection{Comparing Variants of Model Evaluators on Distance (RQ2)}
\label{sec:gnn_eval_dist_exp}
In this experiment, we compare two robust evaluator variants, $\Phi_{eval}^{\text{unif}}$, trained using a uniform edge-drop distribution, and $\Phi_{eval}^{\text{dist}}$, a distance-biased evaluator (see \autoref{sec:evalrob}), which employs a distance bias in edge-dropping, assigning higher removal probabilities to edges that are further away from the head entity. The aim is to determine whether distance-aware edge-dropping produces a more robust evaluator for assessing the quality of localized explanatory subgraphs.

\subsubsection{Experimental Setup} A robust evaluator should preserve predictive performance despite changes in subgraph structure, especially in scenarios involving varying distances from the head entity. Following the proxy strategy from the previous experiment, we vary the radius around the query entity and extract distance-based ego networks for each head entity in our validation set. We then measure each model's MRR at different ego network sizes. Note that the Freebase dataset is dense, and a radius of 4 hops from any node captures a large portion of the graph.

\subsubsection{Results} \autoref{gnn_eval_result_dist} and \autoref{tab:mrr-vs-hop-distance} illustrate the results of this experiment on all four datasets. Both evaluators exhibit comparable accuracy on smaller ego networks, but differences emerge as the subgraph radius increases. On WN18RR, $\Phi_{eval}^{\text{unif}}$ exhibits a counterintuitive performance drop in the range of 3-5 hops despite the addition of new edges, suggesting that the expanded subgraph introduces structural noise that disrupts predictions for some queries. In contrast, the performance of the distance-biased evaluator $\Phi_{eval}^{\text{dist}}$ increases monotonically with larger networks and generally outperforms $\Phi_{eval}^{\text{unif}}$ on all four datasets. These results highlight the utility of using distance-based edge sampling as a training strategy for a perturbation evaluator. This reinforces that explanatory patterns are localized around query entities, and that a locality-aware evaluator provides a more reliable signal for perturbation-based explainers.

\begin{figure}[tb!]
\centering
\ref{sharedlegend1}\\[1pt]
\noindent
\begin{minipage}[t]{0.24\columnwidth}
\centering
\begin{tikzpicture}
\begin{axis}[
    name=plot,
    width=1.2\linewidth,
    height=0.85\linewidth,
    xlabel={Hop distance},
    ylabel={\scriptsize MRR},
    ylabel style={yshift=-2pt},
    xmin=1, xmax=10,
    ymin=0.33, ymax=0.56,
    xtick={1,2,3,4,5,6,7,8,9,10},
    ytick={0.35,0.40,0.45,0.50,0.55},
    tick label style={font=\scriptsize},
    xlabel style={font=\scriptsize,yshift=1ex},
    legend to name=sharedlegend1,
    legend columns=2,
    legend style={
        draw=none,
        font=\scriptsize,
        column sep=0.1cm,
    },
    ymajorgrids=true,
    grid style=dashed,
]
\addplot[
    color=red,
    mark=square,
    mark size=1.5pt
]
coordinates {
    (1,0.348)(2,0.376)(3,0.398)(4,0.369)(5,0.381)
    (6,0.438)(7,0.497)(8,0.535)(9,0.549)(10,0.552)
};
\addplot[
    color={rgb,255:red,0; green,120; blue,130},
    mark=triangle,
    mark size=1.5pt,
]
coordinates {
    (1,0.348)(2,0.382)(3,0.458)(4,0.479)(5,0.485)
    (6,0.501)(7,0.528)(8,0.542)(9,0.549)(10,0.550)
};
\legend{$\Phi_{eval}^{unif}$, $\Phi_{eval}^{dist}$}
\end{axis}
\node[below=1pt, font=\scriptsize\bfseries] at (plot.below south) {(a) WN18RR.};
\end{tikzpicture}
\end{minipage}%
\hfill
\begin{minipage}[t]{0.24\columnwidth}
\centering
\begin{tikzpicture}
\begin{axis}[
    name=plot,
    width=1.2\linewidth,
    height=0.85\linewidth,
    xlabel={Hop distance},
    xmin=1, xmax=4,
    ymin=-0.02, ymax=0.45,
    xtick={1,2,3,4},
    ytick={0.0,0.10,0.20,0.30,0.40},
    tick label style={font=\scriptsize},
    xlabel style={font=\scriptsize,yshift=1ex},
    ymajorgrids=true,
    grid style=dashed,
]
\addplot[
    color=red,
    mark=square,
    mark size=1.5pt
]
coordinates {
    (1,0.005)(2,0.141)(3,0.401)(4,0.422)
};
\addplot[
    color={rgb,255:red,0; green,120; blue,130},
    mark=triangle,
    mark size=1.5pt,
]
coordinates {
    (1,0.005)(2,0.171)(3,0.406)(4,0.419)
};
\end{axis}
\node[below=1pt, font=\scriptsize\bfseries] at (plot.below south) {(b) FB15k-237.};
\end{tikzpicture}
\end{minipage}%
\hfill
\begin{minipage}[t]{0.24\columnwidth}
\centering
\begin{tikzpicture}
\begin{axis}[
    name=plot,
    width=1.2\linewidth,
    height=0.85\linewidth,
    xlabel={Hop distance},
    ylabel style={yshift=-2pt},
    xmin=1, xmax=7,
    ymin=0.20, ymax=0.52,
    xtick={1,2,3,4,5,6,7},
    ytick={0.20,0.25,0.30,0.35,0.40,0.45,0.50},
    tick label style={font=\scriptsize},
    xlabel style={font=\scriptsize,yshift=1ex},
    ymajorgrids=true,
    grid style=dashed,
]
\addplot[
    color=red,
    mark=square,
    mark size=1.5pt
]
coordinates {
    (1,0.269)(2,0.307)(3,0.377)(4,0.435)(5,0.464)(6,0.469)(7,0.470)
};
\addplot[
    color={rgb,255:red,0; green,120; blue,130},
    mark=triangle,
    mark size=1.5pt,
]
coordinates {
    (1,0.277)(2,0.321)(3,0.416)(4,0.469)(5,0.482)(6,0.487)(7,0.500)
};
\end{axis}
\node[below=1pt, font=\scriptsize\bfseries] at (plot.below south) {(c) NELL-995.};
\end{tikzpicture}
\end{minipage}%
\hfill
\begin{minipage}[t]{0.24\columnwidth}
\centering
\begin{tikzpicture}
\begin{axis}[
    name=plot,
    width=1.2\linewidth,
    height=0.85\linewidth,
    xlabel={Hop distance},
    xmin=1, xmax=6,
    ymin=0.25, ymax=0.60,
    xtick={1,2,3,4,5,6},
    ytick={0.25,0.30,0.35,0.40,0.45,0.50,0.55,0.60},
    tick label style={font=\scriptsize},
    xlabel style={font=\scriptsize,yshift=1ex},
    ymajorgrids=true,
    grid style=dashed,
]
\addplot[
    color=red,
    mark=square,
    mark size=1.5pt
]
coordinates {
    (1,0.272)(2,0.333)(3,0.479)(4,0.554)(5,0.570)(6,0.572)
};
\addplot[
    color={rgb,255:red,0; green,120; blue,130},
    mark=triangle,
    mark size=1.5pt,
]
coordinates {
    (1,0.281)(2,0.377)(3,0.496)(4,0.572)(5,0.581)(6,0.582)
};
\end{axis}
\node[below=1pt, font=\scriptsize\bfseries] at (plot.below south) {(d) YAGO3-10.};
\end{tikzpicture}
\end{minipage}%
\caption{Comparing evaluator variants $\Phi_{eval}^{unif}$ and $\Phi_{eval}^{dist}$ on different ego network sizes around the query.}
\label{gnn_eval_result_dist}
\end{figure}

\subsection{Explanations on KG Datasets (RQ3)}
\label{ssec:mainres}

In this section, we evaluate explanation quality following the factual framework introduced in~\autoref{sec:framework}.

\subsubsection{Evaluation}
We evaluate the explanations $\mathcal{G}_S$ under different budget constraints. Specifically, we evaluate explanations with budgets $|\mathcal{G}_S| \leq k$, where $|\cdot|$ denotes the number of edges and $k \in \{25, 50, 75, 100, 300, 500, 1000\}$. This effectively limits the size of the explanation, enhancing interpretability.

\subsubsection{Baselines}
We compare RAW-Explainer against representative baselines from three categories.
\textit{Subgraph-based explainers} learn edge masks without explicit path constraints, including GNN-Explainer~\cite{ying2019gnnexplainer}, a post-hoc method that provides instance-level explanations by optimizing mutual information, and PG-Explainer~\cite{PGExplainer}, a parameterized method that models edge importance distributions across multiple instances.
\textit{Path-based explainers} enforce connectivity through path reasoning, including PaGE-Link~\cite{page-link}, which combines k-core pruning with mask learning, and Power-Link~\cite{powerlink}, which replaces shortest-path search with a graph-powering technique for improved efficiency.
\textit{Structural heuristics} extract subgraphs without learning, including BFS and Random Walk, which traverse the graph from the query entities and serve as simple baselines to assess whether learned explainers capture meaningful structure beyond naive graph proximity.

\subsubsection{Experimental Setup}
All baselines follow the same protocol, using the backbone model NBFNet $\Phi$ and the evaluator $\Phi_{eval}^{dist}$ (see previous section) within the perturbation process. Each method generates explanations for the test queries. The resulting subgraphs are fed back to the backbone model and evaluated using MRR. Hyperparameters for both baselines and our method, along with the best model for each approach, are selected based on average MRR across all budget constraints on the validation set.

\subsubsection{Evaluation Metrics} 
We report MRR as the primary evaluation metric; alternative metrics indicate similar trends. Moreover, since the KG-specific baselines are instance-based, this poses a significant challenge in generating explanations at scale. Hence, we also report the inference time to generate the explanations of the \textit{test} queries.

\begin{figure*}[t!]
  \setlength{\abovecaptionskip}{2pt}
  \centering
  \ref{sharedlegend2}\\[4pt]
\noindent
  \begin{minipage}[t]{0.25\textwidth}
    \centering
    \begin{tikzpicture}
      \begin{axis}[
        name=plot,
        width=1.2\linewidth,
        height=0.85\linewidth,
        xlabel={Edge budget},
        ylabel={\scriptsize MRR},
        ylabel style={yshift=-2pt},
        symbolic x coords={25,50,75,100,300,500,1000},
        xtick=data,
        ymin=0, ymax=0.55,
        scaled y ticks=false,
        tick label style={font=\tiny},
        xlabel style={font=\scriptsize,yshift=1ex},
        legend to name=sharedlegend2,
        legend columns=7,
        legend style={,
          draw=none,,
          font=\scriptsize,,
          column sep=0.1cm,,
        },
        ymajorgrids=true,
        grid style=dashed,
      ]
        \addplot[color=orange, mark=triangle, mark size=1.5pt] coordinates {
          (25,0.384) (50,0.434) (75,0.451) (100,0.462) (300,0.489) (500,0.501) (1000,0.515)
        };
        \addplot[color=black!40, mark=x, mark size=1.5pt] coordinates {
          (25,0.047) (50,0.054) (75,0.058) (100,0.059) (300,0.07) (500,0.074) (1000,0.084)
        };
        \addplot[color=green!60!black, mark=square, mark size=1.5pt] coordinates {
          (25,0.42) (50,0.434) (75,0.436) (100,0.438) (300,0.449) (500,0.452) (1000,0.453)
        };
        \addplot[color=blue!80, mark=diamond, mark size=1.5pt] coordinates {
          (25,0.459) (50,0.459) (75,0.467) (100,0.47) (300,0.476) (500,0.479) (1000,0.483)
        };
        \addplot[color=violet!80, mark=pentagon, mark size=1.5pt] coordinates {
          (25,0.428) (50,0.45) (75,0.445) (100,0.442) (300,0.441) (500,0.441) (1000,0.441)
        };
        \addplot[color=cyan!60!black, mark=otimes, mark size=1.5pt] coordinates {
          (25,0.457) (50,0.453) (75,0.448) (100,0.445) (300,0.439) (500,0.439) (1000,0.439)
        };
        \addplot[thick, color=red!80!black, mark=star, mark size=1.5pt] coordinates {
          (25,0.455) (50,0.489) (75,0.494) (100,0.499) (300,0.509) (500,0.515) (1000,0.521)
        };
        \legend{RW, BFS, GNNExplainer, PGExplainer, PaGELink, PowerLink, RAWExplainer (ours)}
      \end{axis}
      \node[below=1pt, font=\scriptsize\bfseries] at (plot.below south) {(a) WN18RR.};
    \end{tikzpicture}
  \end{minipage}%
  \hspace{4pt}%
  \begin{minipage}[t]{0.25\textwidth}
    \centering
    \begin{tikzpicture}
      \begin{axis}[
        name=plot,
        width=1.2\linewidth,
        height=0.85\linewidth,
        xlabel={Edge budget},
        symbolic x coords={25,50,75,100,300,500,1000},
        xtick=data,
        ymin=0, ymax=0.32,
        scaled y ticks=false,
        tick label style={font=\tiny},
        xlabel style={font=\scriptsize,yshift=1ex},
        ymajorgrids=true,
        grid style=dashed,
      ]
        \addplot[color=orange, mark=triangle, mark size=1.5pt] coordinates {
          (25,0.013) (50,0.025) (75,0.037) (100,0.047) (300,0.098) (500,0.131) (1000,0.18)
        };
        \addplot[color=black!40, mark=x, mark size=1.5pt] coordinates {
          (25,0.005) (50,0.005) (75,0.005) (100,0.006) (300,0.01) (500,0.013) (1000,0.016)
        };
        \addplot[color=green!60!black, mark=square, mark size=1.5pt] coordinates {
          (25,0.156) (50,0.174) (75,0.182) (100,0.186) (300,0.177) (500,0.162) (1000,0.139)
        };
        \addplot[color=blue!80, mark=diamond, mark size=1.5pt] coordinates {
          (25,0.165) (50,0.192) (75,0.205) (100,0.212) (300,0.221) (500,0.216) (1000,0.199)
        };
        \addplot[color=violet!80, mark=pentagon, mark size=1.5pt] coordinates {
          (25,0.149) (50,0.15) (75,0.151) (100,0.151) (300,0.151) (500,0.151) (1000,0.151)
        };
        \addplot[color=cyan!60!black, mark=otimes, mark size=1.5pt] coordinates {
          (25,0.154) (50,0.155) (75,0.156) (100,0.157) (300,0.157) (500,0.157) (1000,0.157)
        };
        \addplot[thick, color=red!80!black, mark=star, mark size=1.5pt] coordinates {
          (25,0.117) (50,0.188) (75,0.217) (100,0.234) (300,0.275) (500,0.287) (1000,0.299)
        };
      \end{axis}
      \node[below=1pt, font=\scriptsize\bfseries] at (plot.below south) {(b) FB15k-237.};
    \end{tikzpicture}
  \end{minipage}%
  \hspace{-4pt}%
  \begin{minipage}[t]{0.25\textwidth}
    \centering
    \begin{tikzpicture}
      \begin{axis}[
        name=plot,
        width=1.2\linewidth,
        height=0.85\linewidth,
        xlabel={Edge budget},
        symbolic x coords={25,50,75,100,300,500,1000},
        xtick=data,
        ymin=0, ymax=0.45,
        scaled y ticks=false,
        tick label style={font=\tiny},
        xlabel style={font=\scriptsize,yshift=1ex},
        ymajorgrids=true,
        grid style=dashed,
      ]
        \addplot[color=orange, mark=triangle, mark size=1.5pt] coordinates {
          (25,0.283) (50,0.343) (75,0.378) (100,0.394) (300,0.413) (500,0.404) (1000,0.422)
        };
        \addplot[color=black!40, mark=x, mark size=1.5pt] coordinates {
          (25,0.02) (50,0.021) (75,0.028) (100,0.029) (300,0.063) (500,0.066) (1000,0.073)
        };
        \addplot[color=green!60!black, mark=square, mark size=1.5pt] coordinates {
          (25,0.316) (50,0.318) (75,0.314) (100,0.306) (300,0.257) (500,0.218) (1000,0.178)
        };
        \addplot[color=blue!80, mark=diamond, mark size=1.5pt] coordinates {
          (25,0.277) (50,0.291) (75,0.305) (100,0.305) (300,0.303) (500,0.283) (1000,0.284)
        };
        \addplot[color=violet!80, mark=pentagon, mark size=1.5pt] coordinates {
          (25,0.294) (50,0.298) (75,0.297) (100,0.298) (300,0.297) (500,0.298) (1000,0.298)
        };
        \addplot[color=cyan!60!black, mark=otimes, mark size=1.5pt] coordinates {
          (25,0.254) (50,0.256) (75,0.256) (100,0.256) (300,0.256) (500,0.256) (1000,0.256)
        };
        \addplot[thick, color=red!80!black, mark=star, mark size=1.5pt] coordinates {
          (25,0.340) (50,0.367) (75,0.382) (100,0.392) (300,0.425) (500,0.427) (1000,0.438)
        };
      \end{axis}
      \node[below=1pt, font=\scriptsize\bfseries] at (plot.below south) {(c) NELL-995.};
    \end{tikzpicture}
  \end{minipage}%
  \hspace{0pt}%
  \begin{minipage}[t]{0.25\textwidth}
    \centering
    \begin{tikzpicture}
      \begin{axis}[
        name=plot,
        width=1.2\linewidth,
        height=0.85\linewidth,
        xlabel={Edge budget},
        symbolic x coords={25,50,75,100,300,500,1000},
        xtick=data,
        ymin=0, ymax=0.53,
        scaled y ticks=false,
        tick label style={font=\tiny},
        xlabel style={font=\scriptsize,yshift=1ex},
        ymajorgrids=true,
        grid style=dashed,
      ]
        \addplot[color=orange, mark=triangle, mark size=1.5pt] coordinates {
          (25,0.202) (50,0.295) (75,0.326) (100,0.345) (300,0.403) (500,0.43) (1000,0.463)
        };
        \addplot[color=black!40, mark=x, mark size=1.5pt] coordinates {
          (25,0.014) (50,0.018) (75,0.046) (100,0.047) (300,0.052) (500,0.059) (1000,0.071)
        };
        \addplot[color=green!60!black, mark=square, mark size=1.5pt] coordinates {
          (25,0.291) (50,0.322) (75,0.338) (100,0.341) (300,0.341) (500,0.338) (1000,0.326)
        };
        \addplot[color=blue!80, mark=diamond, mark size=1.5pt] coordinates {
          (25,0.234) (50,0.283) (75,0.313) (100,0.329) (300,0.386) (500,0.403) (1000,0.413)
        };
        \addplot[color=violet!80, mark=pentagon, mark size=1.5pt] coordinates {
          (25,0.308) (50,0.308) (75,0.308) (100,0.308) (300,0.308) (500,0.308) (1000,0.308)
        };
        \addplot[color=cyan!60!black, mark=otimes, mark size=1.5pt] coordinates {
          (25,0.245) (50,0.244) (75,0.243) (100,0.243) (300,0.243) (500,0.243) (1000,0.243)
        };
        \addplot[thick, color=red!80!black, mark=star, mark size=1.5pt] coordinates {
          (25,0.401) (50,0.424) (75,0.447) (100,0.459) (300,0.487) (500,0.495) (1000,0.504)
        };
      \end{axis}
      \node[below=1pt, font=\scriptsize\bfseries] at (plot.below south) {(d) YAGO3-10.};
    \end{tikzpicture}
  \end{minipage}%
  \caption{Test MRR vs.\ edge budget for each dataset.}
  \label{fig:mrr-vs-edgebudget}
\end{figure*}

\begin{table}[!ht]
\centering
\footnotesize
\caption{Inference time (minutes) to generate test-set explanations. \textbf{Best} and \underline{second best} are highlighted.}
\label{tab:inference_time}
\setlength{\tabcolsep}{0pt}
\begin{tabular*}{\columnwidth}{@{\extracolsep{\fill}}lcccc}
\toprule
Method & WN18RR & FB15k-237 & NELL-995 & YAGO3-10 \\ 
\midrule
GNN-Explainer & 109 & 1663 & 90 & 429 \\
PG-Explainer & \textbf{6} & \textbf{48} & \textbf{4} & \textbf{28} \\
PaGE-Link & 172 & 940 & 700 & 2238 \\
Power-Link & 166 & 9577 & 127 & 183 \\ 
\midrule
\textbf{RAW-Explainer} & \underline{17} & \underline{128} & \underline{14} & \underline{91} \\ 
\bottomrule
\end{tabular*}
\end{table}
\subsubsection{Results}
\autoref{fig:mrr-vs-edgebudget} presents the MRR results across all four datasets (see \autoref{detailed-mrr}). Tables~\ref{tab:inference_time} and~\ref{tab:connectivity_wn18rr} report inference time and connectivity, respectively.

\paragraph{Explanation quality.}
RAW-Explainer outperforms all baselines across most budget levels on all datasets and shows steady improvement as the edge budget increases, demonstrating its ability to progressively integrate informative edges into the explanations. On FB15k-237 in particular, RAW-Explainer starts below PG-Explainer at the smallest budgets (25 and 50 edges) but overtakes it from 75 edges onward, with the gap widening substantially at larger budgets. We attribute this initial lag to the high density of the Freebase graph. At very small budgets, the random walk has limited capacity to distinguish important edges among many candidates. As the budget increases, however, the selected edges begin to form coherent multi-hop patterns, making RAW-Explainer's advantage clear. These results highlight RAW-Explainer's strategy: it trades edge selection precision for connectivity at low budgets, but this limitation is quickly mitigated as larger edge budgets are introduced.

\paragraph{Structural baselines.}
The Random Walk baseline performs surprisingly well, particularly on WN18RR and NELL-995, where it is competitive with or outperforms several learned explainers. This observation aligns with recent work~\cite{DBLP:conf/iclr/ZhouZYY024}, which demonstrates the effectiveness of random walks for edge selection in knowledge graphs. Although not a model-based explainer, this result underscores the importance of connectivity and locality in knowledge graph explanations and supports our design choice of incorporating random walks as a core component of RAW-Explainer rather than relying solely on learned edge masks. BFS, in contrast, consistently ranks last across all datasets. A fully random edge selection baseline (not shown) yields near-zero MRR, confirming that even simple proximity-based traversal captures meaningful structure compared to uniformly random selection.

\paragraph{Learned baselines.}
Among the learned baselines, PG-Explainer is the most consistently competitive. GNN-Explainer shows strong performance at small subgraph sizes, particularly on NELL-995 and YAGO3-10, which we attribute to its instance-level optimization that favors compact explanations. However, its performance generally plateaus or declines at higher budgets, indicating difficulty in scaling to larger explanations. PaGE-Link and Power-Link exhibit highly correlated behavior, which is expected given their shared reliance on shortest-path-based reasoning. PaGE-Link performs competitively at smaller budgets on WN18RR, suggesting effectiveness on simpler relational structures, but both methods plateau quickly on denser graphs.

\paragraph{Inference time and scalability.}
\label{inf-time}

\autoref{tab:inference_time} summarizes the inference time (in minutes) for generating test-set explanations across all methods. PG-Explainer is the fastest method on all four datasets, followed by RAW-Explainer. On FB15k-237, Power-Link requires over 9500 minutes, compared to 128 minutes for RAW-Explainer. PaGE-Link and Power-Link also face scalability limits during training. On YAGO3-10, both methods could only be trained on a 20\% subset of the data due to GPU memory constraints, with Power-Link further restricted to a maximum path length of one because of rapid memory growth inherent to its architecture.

\paragraph{Connectivity analysis.}
\label{connectivity}

We report the average number of connected components per explanation across varying edge budgets on the WN18RR dataset for RAW-Explainer and PG-Explainer in Table~\ref{tab:connectivity_wn18rr}. PG-Explainer struggles to maintain connectivity as the explanation size grows; its average number of connected components drastically increases alongside the edge budget. In contrast, RAW-Explainer maintains a consistently low and stable number of connected components across all edge budgets. These results show that RAW-Explainer produces less fragmented explanations than PG-Explainer, consistent with its connectivity objective. The qualitative case study in \autoref{case-study} illustrates these structural differences.

\begin{table}[ht!]
\centering
\caption{Average number of connected components on WN18RR across varying edge budgets.}
\label{tab:connectivity_wn18rr}
\begin{tabular}{l ccc ccc c}
\toprule
\textbf{Method} & \textbf{25} & \textbf{50} & \textbf{75} & \textbf{100} & \textbf{300} & \textbf{500} & \textbf{1000} \\
\midrule
PGExplainer  & 1.49 & 1.57 & 1.68 & 1.80 & 2.73 & 3.74 & 8.66 \\
RAWExplainer & \textbf{1.39} & \textbf{1.25} & \textbf{1.21} & \textbf{1.19} & \textbf{1.22} & \textbf{1.28} & \textbf{1.39} \\
\bottomrule
\end{tabular}
\end{table}

Ablation studies on WN18RR are reported in \autoref{app:ablation}. Further evaluations cover backbone sensitivity (\autoref{app:backbone-sensitivity}) and alternative explanation objectives (\autoref{app:other-objectives}).

\section{Conclusions}
\label{conclusion}
This paper presents RAW-Explainer, a novel explainability framework for path reasoning GNNs that generates connected and interpretable subgraph explanations for link prediction in knowledge graphs. By using a neural network to parameterize the generation of a learned edge mask and coupling it with graph powering techniques for generating random walks, RAW-Explainer encourages connectivity while supporting efficient generation of explanations across queries. To overcome the distribution shift between the full knowledge graph and the much smaller explanatory subgraphs, we integrate a novel distance-based robust evaluator that generalizes effectively to the subgraph distribution. Extensive experiments demonstrate that RAW-Explainer strikes a balance between generating concise and connected subgraphs that enhance interpretability, effectiveness, and efficiency compared to existing methods.

\section*{Acknowledgments}
We thank the anonymous reviewers for their feedback and constructive discussion during the rebuttal period. This research was carried out on the High Performance Computing resources at New York University Abu Dhabi.

\bibliographystyle{unsrtnat}
\bibliography{chapters/8_reference}
\clearpage
\appendix
\section{Datasets}
We evaluate path-based GNN explainability methods on four standard KG completion benchmarks. \autoref{tab:dataset-stats} summarizes their statistics.\begin{table}[!ht]
\centering
\footnotesize
\caption{Dataset statistics for KG completion.}
\label{tab:dataset-stats}
\begin{tabular}{lcccc}
\toprule
 & \textbf{FB15k-237} & \textbf{WN18RR} & \textbf{NELL-995} & \textbf{YAGO3-10} \\ 
\midrule
\#Entity     & 14,541  & 40,943 & 75,492  & 123,182 \\
\#Relation   & 237     & 11     & 200     & 37 \\
\#Train      & 272,115 & 86,835 & 149,678 & 1,079,040 \\
\#Validation & 17,535  & 3,034  & 543     & 5,000 \\
\#Test       & 20,466  & 3,134  & 2,818   & 5,000 \\ 
\bottomrule
\end{tabular}
\end{table}

\section{Backbone GNN and Evaluator Configurations}
\label{appendix}
We provide the configuration of the backbone GNN $\Phi$ and evaluators $\Phi_{eval}^{\text{unif}}, \Phi_{eval}^{\text{dist}}$ used in the NBFNet experiments in \autoref{tab:phi-params}. These models share the following architecture and optimization settings: NBFNet~\cite{nbfnet} with 6 layers, 32-dimensional embeddings, a DistMult~\cite{DBLP:journals/corr/YangYHGD14a} message function to score weighted messages along edges, Principal Neighbourhood Aggregation (PNA) for neighborhood messages, shortcut connections to enable residual connections between GNN layers, layer normalization, and the Adam optimizer with a learning rate $1\times10^{-3}$. \texttt{Dep} (Dependent Parameter) specifies whether relation embeddings are query-dependent across layers, \texttt{1-hop} (Remove One-Hop) controls the removal of direct head–tail edges to prevent shortcut predictions, and \texttt{DD} denotes Distance Dropout. For further details about NBFNet's default configurations, please refer to \cite{nbfnet}.

\begin{table}[!ht]
\centering
\footnotesize
\caption{Evaluator-specific hyperparameters across datasets.}
\label{tab:phi-params}
\setlength{\tabcolsep}{4pt}
\begin{tabular}{llcccccc}
\toprule
Model & Data & Dep & 1-hop & $p_e$ & DD & dist & $p_{\max}$ \\ 
\midrule
$\Phi$
& WN18RR    & N & N & 0.0 & N & 0  & 0.0 \\
& FB15k-237 & Y & Y & 0.0 & N & 0  & 0.0 \\
& NELL-995  & N & N & 0.0 & N & 0  & 0.0 \\
& YAGO3-10  & N & N & 0.0 & N & 0  & 0.0 \\ \hline

$\Phi_{\text{eval}}^{\text{unif}}$
& WN18RR    & N & N & 0.15 & N & 0  & 0.0 \\
& FB15k-237 & Y & Y & 0.15 & N & 0  & 0.0 \\
& NELL-995  & N & N & 0.20 & N & 0  & 0.0 \\
& YAGO3-10  & N & N & 0.10 & N & 0  & 0.0 \\ \hline

$\Phi_{\text{eval}}^{\text{dist}}$
& WN18RR    & N & N & 0.04 & Y & 10 & 0.10 \\
& FB15k-237 & Y & Y & 0.05 & Y & 4  & 0.20 \\
& NELL-995  & N & N & 0.05 & Y & 10 & 0.10 \\
& YAGO3-10  & N & N & 0.04 & Y & 10 & 0.10 \\ \hline
\end{tabular}
\end{table}

For $\Phi_{eval}^{\text{dist}}$, the edge dropout probability is $p_{\text{drop}} = \min(p_e \cdot \text{distance},\; p_{\max})$, where $p_e$ is the base edge dropout probability, $\text{distance}$ is the edge's hop distance from the query entity (starting at 1 for incident edges), and $p_{\max}$ caps the dropout probability. The \texttt{dist} hyperparameter limits the hop search used to compute edge distances; edges not reached within this search are dropped with probability $p_{\max}$.

\section{Explainer Configurations}
This section reports explainer-specific hyperparameters used across all experiments to ensure fair and reproducible comparisons. \autoref{tab:shared-params}–\autoref{tab:powerlink-params} summarize the model-specific hyperparameters for each explainer evaluated in this work. Models denoted by $\dagger$ were trained on a 20\% subset of the YAGO3-10 dataset due to GPU memory constraints; for these models, experiments were repeated across multiple random seeds, and the best results are reported. \\

\subsection{Random Walk and BFS}
The hyperparameters for Random Walk and BFS strategies are listed in \autoref{tab:shared-params}. The \emph{Top-k tails} $=0$ setting restricts the set of relevant entities $\rho$ to the query head $h$. The \emph{Ego network} flag indicates whether the subgraph construction is restricted to the ego network around the set of relevant entities $\rho$. This ego network is constructed to match the computational graph used in the backbone GNN $\Phi$ for the final representation of the entities in $\rho$, following \cite{ying2019gnnexplainer, PGExplainer}.

\begin{table}[!ht]
\centering
\footnotesize
\caption{Hyperparameters used across all datasets.}
\label{tab:shared-params}

\begin{minipage}[t]{0.48\linewidth}
\centering
\textbf{RandomWalk}\par\vspace{4pt}
\begin{tabular}{ll}
\toprule
Parameter & Value \\
\midrule
Top-$k$ tails     & 0 \\
Teleport prob.    & 0.2 \\
Adj. aggregation  & max \\
Ego network       & No \\
\bottomrule
\end{tabular}
\end{minipage}\hfill
\begin{minipage}[t]{0.48\linewidth}
\centering
\textbf{BFS}\par\vspace{4pt}
\begin{tabular}{ll}
\toprule
Parameter & Value \\
\midrule
Top-$k$ tails & 0 \\
Ego network   & No \\
\bottomrule
\end{tabular}
\end{minipage}
\end{table}

\subsection{GNN-Explainer}
\autoref{tab:gnnexplainer-params} summarizes the model hyperparameters for GNN-Explainer. These include optimization settings (\emph{optimizer} and \emph{learning rate}) and regularization coefficients that control explanation size and stability.

\begin{table}[!ht]
\centering
\footnotesize
\caption{Hyperparameters used for GNN-Explainer on each dataset.}
\label{tab:gnnexplainer-params}
\setlength{\tabcolsep}{6pt}
\begin{tabular}{lcccc}
\toprule
Parameter & WN18RR & FB15k-237 & NELL-995 & YAGO3-10 \\
\midrule
Optimizer              & Adam & Adam & Adam & Adam \\
Learning rate          & $1\times10^{-2}$ & $1\times10^{-3}$ & $1\times10^{-2}$ & $1\times10^{-2}$ \\
Size regularization    & 1 & 4 & 1 & 1 \\
Entropy regularization & 1 & 3 & 1 & 1 \\
Top-$k$ tails          & 1 & 1 & 1 & 1 \\
Ego network            & No & No & No & No \\ \hline
\end{tabular}
\end{table}

\subsection{PG-Explainer}
\autoref{tab:pgexplainer-params} highlights the hyperparameters used for PG-Explainer. In addition to optimization and architecture settings, \emph{Temperature (start)} and \emph{Temperature (end)} control the stochasticity of edge sampling during training, whereas the \emph{Sample bias} parameter determines the range of epsilon in the reparameterization trick.

\begin{table}[!ht]
\centering
\footnotesize
\caption{Hyperparameters used for PG-Explainer on each dataset.}
\label{tab:pgexplainer-params}
\setlength{\tabcolsep}{6pt}
\begin{tabular}{lcccc}
\toprule
Parameter & WN18RR & FB15k-237 & NELL-995 & YAGO3-10 \\ \hline
Optimizer              & Adam & Adam & Adam & Adam \\
Learning rate          & $1\times10^{-3}$ & $5\times10^{-5}$ & $1\times10^{-3}$ & $5\times10^{-5}$ \\
Size regularization    & 2 & 5 & 2 & 4 \\
Entropy regularization & 4 & 2 & 4 & 2 \\
Temp. (start)          & 4 & 2 & 4 & 2 \\
Temp. (end)            & 3 & 2 & 3 & 2 \\
Sample bias            & 0.5 & 0.5 & 0.5 & 0.4 \\
\#MLP layers           & 3 & 3 & 3 & 2 \\
Top-$k$ tails          & 1 & 1 & 1 & 1 \\
Ego network            & Yes & Yes & Yes & Yes \\ \hline
\end{tabular}
\end{table}

\subsection{RAW-Explainer}
As shown in \autoref{tab:rawexplainer-params}, RAW-Explainer extends prior hyperparameter configurations with a random-walk loss controlled by the boolean flag \emph{Random walk loss}. Additional parameters regulate random-walk traversal and regularization, including teleport behavior (\emph{Teleport probability}), node selection (\emph{RW top-k nodes}), and weighting of components inside and outside the subgraph (\emph{RW inside loss weight} and \emph{RW outside loss weight}, respectively). Finally, edges with mask probabilities below the \emph{Threshold} are filtered out.

\begin{table}[!ht]
\centering
\footnotesize
\caption{Hyperparameters used for RAW-Explainer on each dataset.}
\label{tab:rawexplainer-params}
\setlength{\tabcolsep}{6pt}
\begin{tabular}{lcccc}
\toprule
Parameter & WN18RR & FB15k-237 & NELL-995 & YAGO3-10 \\ \hline
Optimizer & Adam & AdamW & Adam & Adam \\
Learning rate & $5\times10^{-5}$ & $3\times10^{-1}$ & $5\times10^{-5}$ & $5\times10^{-5}$ \\
LR decay & - & $5\times10^{-3}$ & - & - \\
Size reg. & 4 & 1 & 4 & 4 \\
Entropy reg. & 2 & 4 & 2 & 2 \\
Temp. (start) & 2 & 2 & 2 & 2 \\
Temp. (end) & 2 & 1 & 2 & 2 \\
Sample bias & 0.4 & 0.5 & 0.4 & 0.4 \\
Rel. emb. & Yes & Yes & Yes & Yes \\
\#MLP layers & 2 & 5 & 2 & 2 \\
Top-$k$ tails & 2 & 4 & 2 & 2 \\
Ego net. & No & No & No & No \\
Threshold & 0 & 0.5 & 0 & 0 \\ \hline

RW loss & Yes & Yes & Yes & Yes \\
Adj. agg. & sum & sum & sum & sum \\
Teleport prob. & 0.2 & 0.5 & 0.2 & 0.2 \\
RW top-$k$ nodes & 500 & 250 & 500 & 500 \\
RW in loss wt. & 1.0 & 1.5 & 1.0 & 1.0 \\
RW out loss wt. & 0.5 & 0.75 & 0.5 & 0.5 \\
Teleport adj. & Yes & Yes & Yes & Yes \\ \hline
\end{tabular}
\end{table}

\subsection{PaGE-Link}
\autoref{tab:pagelink-params} reports the hyperparameters used for PaGE-Link. With the \emph{Prune graph} flag enabled, graph pruning is performed through degree (\emph{Prune max degree}) and k-core–based (\emph{k-core}) filtering. PaGE-Link further uses a path-based loss that penalizes edges off paths during training, with coefficients $\alpha$ and $\beta$ weighting on-path and off-path edges, respectively. The \emph{\#Paths} parameter limits the number of shortest paths considered in this loss. In addition, all datasets use a \emph{Timeout duration} of 3 seconds during path finding to prevent excessive computation.

\begin{table}[!ht]
\centering
\footnotesize
\caption{Hyperparameters used for PaGE-Link on each dataset.}
\label{tab:pagelink-params}
\setlength{\tabcolsep}{6pt}
\begin{tabular}{lcccc}
\toprule
Parameter & WN18RR & FB15k-237 & NELL-995 & YAGO3-10$\dagger$ \\ \hline
Optimizer            & Adam & Adam & Adam & Adam \\
Learning rate        & $1\times10^{-2}$ & $1\times10^{-2}$ & $1\times10^{-2}$ & $1\times10^{-2}$ \\
Prune graph          & Yes & Yes & Yes & Yes \\
Path loss            & Yes & Yes & Yes & Yes \\
Top-$k$ tails        & 1   & 1   & 1   & 1 \\
Prune max deg.       & 200 & 30  & 200 & 200 \\
$k$-core             & 2   & 3   & 2   & 2 \\
\#Paths              & 4   & 2   & 4   & 4 \\
$\alpha$             & 1.0 & 0.5 & 1.0 & 1.0 \\
$\beta$              & 1.5 & 0.5 & 1.5 & 1.5 \\
Ego network          & Yes & Yes & Yes & Yes \\
Timeout (s)          & 3.0 & 3.0 & 3.0 & 3.0 \\ \hline
\end{tabular}
\end{table}

\subsection{Power-Link}
Power-Link follows a hyperparameter configuration similar to that of PaGE-Link, with \emph{Adjacency aggregation} controlling multigraph-to-digraph conversion for path computation and \emph{Max path length} limiting the path length used in path loss.

\begin{table}[!ht]
\centering
\footnotesize
\caption{Hyperparameters used for Power-Link on each dataset.}
\label{tab:powerlink-params}
\setlength{\tabcolsep}{6pt}
\begin{tabular}{lcccc}
\toprule
Parameter & WN18RR & FB15k-237 & NELL-995 & YAGO3-10$\dagger$ \\ \hline
Optimizer           & Adam & Adam & Adam & Adam \\
Learning rate       & $5\times10^{-3}$ & $5\times10^{-3}$ & $5\times10^{-3}$ & $5\times10^{-3}$ \\
Prune graph         & Yes & Yes & Yes & Yes \\
Prune max deg.      & 20 & 30 & 20 & 30 \\
Path loss           & Yes & Yes & Yes & Yes \\
Top-$k$ tails       & 1 & 1 & 1 & 1 \\
$k$-core            & 2 & 2 & 2 & 2 \\
\#MLP layers        & 1 & 1 & 1 & 1 \\
Adj. aggregation    & max & max & max & max \\
Max path len.       & 3 & 3 & 3 & 1 \\
Size reg.           & 1 & 1 & 1 & 1 \\
Ego network         & Yes & Yes & Yes & Yes \\
Timeout (s)         & 3.0 & 3.0 & 3.0 & 3.0 \\ \hline
\end{tabular}
\end{table}

\section{Robustness to Random Edge Dropping}
\label{random-edge-drop}

\autoref{tab:mrr-vs-random-drop} summarizes the MRR under different random edge-drop probabilities for $\Phi$ and $\Phi_{\mathrm{eval}}$.

\begin{table}[!ht]
\centering
\footnotesize
\caption{MRR under different random edge-drop probabilities.}
\label{tab:mrr-vs-random-drop}
\setlength{\tabcolsep}{2pt}
\begin{tabular*}{\columnwidth}{@{\extracolsep{\fill}}lcccccccc}
\toprule
& \multicolumn{2}{c}{WN18RR}
& \multicolumn{2}{c}{FB15k-237}
& \multicolumn{2}{c}{NELL-995}
& \multicolumn{2}{c}{YAGO3-10} \\
\cmidrule(lr){2-3}
\cmidrule(lr){4-5}
\cmidrule(lr){6-7}
\cmidrule(lr){8-9}
Drop Prob.
& $\Phi$ & $\Phi_{\mathrm{eval}}$
& $\Phi$ & $\Phi_{\mathrm{eval}}$
& $\Phi$ & $\Phi_{\mathrm{eval}}$
& $\Phi$ & $\Phi_{\mathrm{eval}}$ \\
\midrule
0.00 & 0.551 & 0.553 & 0.422 & 0.427 & 0.462 & 0.472 & 0.578 & 0.572 \\
0.05 & 0.488 & 0.524 & 0.411 & 0.419 & --    & --    & --    & --    \\
0.10 & 0.437 & 0.496 & 0.402 & 0.412 & 0.130 & 0.455 & 0.514 & 0.532 \\
0.20 & 0.353 & 0.444 & 0.383 & 0.396 & 0.099 & 0.436 & 0.464 & 0.489 \\
0.30 & 0.288 & 0.389 & 0.361 & 0.382 & 0.087 & 0.414 & 0.407 & 0.437 \\
0.40 & 0.236 & 0.341 & 0.337 & 0.363 & 0.072 & 0.390 & 0.346 & 0.397 \\
0.50 & 0.196 & 0.289 & 0.310 & 0.342 & 0.057 & 0.364 & 0.283 & 0.349 \\
\bottomrule
\end{tabular*}
\end{table}

\section{Effect of Hop Distance}
\label{hop-distance}

\autoref{tab:mrr-vs-hop-distance} summarizes the MRR at different hop distances for $\Phi_{\mathrm{eval}}^{\mathrm{unif}}$ and $\Phi_{\mathrm{eval}}^{\mathrm{dist}}$.

\begin{table}[!ht]
\centering
\footnotesize
\caption{MRR at different hop distances.}
\label{tab:mrr-vs-hop-distance}
\setlength{\tabcolsep}{2pt}
\begin{tabular*}{\columnwidth}{@{\extracolsep{\fill}}lcccccccc}
\toprule
& \multicolumn{2}{c}{WN18RR}
& \multicolumn{2}{c}{FB15k-237}
& \multicolumn{2}{c}{NELL-995}
& \multicolumn{2}{c}{YAGO3-10} \\
\cmidrule(lr){2-3}
\cmidrule(lr){4-5}
\cmidrule(lr){6-7}
\cmidrule(lr){8-9}
Hops
& $\Phi_{\mathrm{eval}}^{\mathrm{unif}}$
& $\Phi_{\mathrm{eval}}^{\mathrm{dist}}$
& $\Phi_{\mathrm{eval}}^{\mathrm{unif}}$
& $\Phi_{\mathrm{eval}}^{\mathrm{dist}}$
& $\Phi_{\mathrm{eval}}^{\mathrm{unif}}$
& $\Phi_{\mathrm{eval}}^{\mathrm{dist}}$
& $\Phi_{\mathrm{eval}}^{\mathrm{unif}}$
& $\Phi_{\mathrm{eval}}^{\mathrm{dist}}$ \\
\midrule
1  & 0.348 & 0.348 & 0.005 & 0.005 & 0.269 & 0.277 & 0.272 & 0.281 \\
2  & 0.376 & 0.382 & 0.141 & 0.171 & 0.307 & 0.321 & 0.333 & 0.377 \\
3  & 0.398 & 0.458 & 0.401 & 0.406 & 0.377 & 0.416 & 0.479 & 0.496 \\
4  & 0.369 & 0.479 & 0.422 & 0.419 & 0.435 & 0.469 & 0.554 & 0.572 \\
5  & 0.381 & 0.485 & --    & --    & 0.464 & 0.482 & 0.570 & 0.581 \\
6  & 0.438 & 0.501 & --    & --    & 0.469 & 0.487 & 0.572 & 0.582 \\
7  & 0.497 & 0.528 & --    & --    & 0.470 & 0.500 & --    & --    \\
8  & 0.535 & 0.542 & --    & --    & --    & --    & --    & --    \\
9  & 0.549 & 0.549 & --    & --    & --    & --    & --    & --    \\
10 & 0.552 & 0.550 & --    & --    & --    & --    & --    & --    \\
\bottomrule
\end{tabular*}
\end{table}

\section{Detailed MRR Results Across Edge Budgets}
\autoref{tab:detailed-mrr} reports the test MRR of all methods across edge budgets on the four datasets. These results provide the detailed values corresponding to the trends shown in \autoref{fig:mrr-vs-edgebudget}.

\label{detailed-mrr}
\begin{table*}[htbp]
\centering
\footnotesize
\caption{Test MRR across different edge budgets on all four datasets. 
\textbf{Best} and \underline{second best} results are highlighted.}
\label{tab:detailed-mrr}
\setlength{\tabcolsep}{3pt}
\begin{tabular*}{\textwidth}{@{\extracolsep{\fill}}lccccccc}
\toprule
Method & 25 & 50 & 75 & 100 & 300 & 500 & 1000 \\
\midrule
\multicolumn{8}{c}{\textbf{WN18RR}} \\
\midrule
\textit{RW}             & 0.384 & 0.434 & 0.451 & 0.462 & \underline{0.489} & \underline{0.501} & \underline{0.515} \\
\textit{BFS}            & 0.047 & 0.054 & 0.058 & 0.059 & 0.070 & 0.074 & 0.084 \\
GNN-Explainer           & 0.420 & 0.434 & 0.436 & 0.438 & 0.449 & 0.452 & 0.453 \\
PG-Explainer            & \textbf{0.459} & \underline{0.459} & \underline{0.467} & \underline{0.470} & 0.476 & 0.479 & 0.483 \\
PaGE-Link               & 0.428 & 0.450 & 0.445 & 0.442 & 0.441 & 0.441 & 0.441 \\
Power-Link              & \underline{0.457} & 0.453 & 0.448 & 0.445 & 0.439 & 0.439 & 0.439 \\
\textbf{RAW-Explainer}  & 0.455 & \textbf{0.489} & \textbf{0.494} & \textbf{0.499}
                        & \textbf{0.509} & \textbf{0.515} & \textbf{0.521} \\
\midrule
\multicolumn{8}{c}{\textbf{FB15k-237}} \\
\midrule
\textit{RW}             & 0.013 & 0.025 & 0.037 & 0.047 & 0.098 & 0.131 & 0.180 \\
\textit{BFS}            & 0.005 & 0.005 & 0.005 & 0.006 & 0.010 & 0.013 & 0.016 \\
GNN-Explainer           & \underline{0.156} & 0.174 & 0.182 & 0.186 & 0.177 & 0.162 & 0.139 \\
PG-Explainer            & \textbf{0.165} & \textbf{0.192} & \underline{0.205}
                        & \underline{0.212} & \underline{0.221} & \underline{0.216}
                        & \underline{0.199} \\
PaGE-Link               & 0.149 & 0.150 & 0.151 & 0.151 & 0.151 & 0.151 & 0.151 \\
Power-Link              & 0.154 & 0.155 & 0.156 & 0.157 & 0.157 & 0.157 & 0.157 \\
\textbf{RAW-Explainer}  & 0.117 & \underline{0.188} & \textbf{0.217} & \textbf{0.234}
                        & \textbf{0.275} & \textbf{0.287} & \textbf{0.299} \\
\midrule
\multicolumn{8}{c}{\textbf{NELL-995}} \\
\midrule
\textit{RW}             & 0.283 & \underline{0.343} & \underline{0.378} & \textbf{0.394} & \underline{0.413} & \underline{0.404} & \underline{0.422} \\
\textit{BFS}            & 0.020 & 0.021 & 0.028 & 0.029 & 0.063 & 0.066 & 0.073 \\
GNN-Explainer           & \underline{0.316} & 0.318 & 0.314 & 0.306 & 0.257 & 0.218 & 0.178 \\
PG-Explainer            & 0.277 & 0.291 & 0.305 & 0.305 & 0.303 & 0.283 & 0.284 \\
PaGE-Link               & 0.294 & 0.298 & 0.297 & 0.298 & 0.297 & 0.298 & 0.298 \\
Power-Link              & 0.254 & 0.256 & 0.256 & 0.256 & 0.256 & 0.256 & 0.256 \\
\textbf{RAW-Explainer}  & \textbf{0.340} & \textbf{0.367} & \textbf{0.382}
                        & \underline{0.392} & \textbf{0.425} & \textbf{0.427}
                        & \textbf{0.438} \\
\midrule
\multicolumn{8}{c}{\textbf{YAGO3-10}} \\
\midrule
\textit{RW}             & 0.202 & 0.295 & 0.326 & \underline{0.345} & \underline{0.403}
                        & \underline{0.430} & \underline{0.463} \\
\textit{BFS}            & 0.014 & 0.018 & 0.046 & 0.047 & 0.052 & 0.059 & 0.071 \\
GNN-Explainer           & 0.291 & \underline{0.322} & \underline{0.338} & 0.341
                        & 0.341 & 0.338 & 0.326 \\
PG-Explainer            & 0.234 & 0.283 & 0.313 & 0.329 & 0.386 & 0.403 & 0.413 \\
PaGE-Link               & \underline{0.308} & 0.308 & 0.308 & 0.308 & 0.308 & 0.308 & 0.308 \\
Power-Link              & 0.245 & 0.244 & 0.243 & 0.243 & 0.243 & 0.243 & 0.243 \\
\textbf{RAW-Explainer}  & \textbf{0.401} & \textbf{0.424} & \textbf{0.447}
                        & \textbf{0.459} & \textbf{0.487} & \textbf{0.495}
                        & \textbf{0.504} \\
\bottomrule
\end{tabular*}
\end{table*}

\section{Ablation Study}
\label{app:ablation}

We conduct ablation studies to systematically validate our central claims on WN18RR. Table~\ref{tab:ablation-wn18rr} reports the results across varying edge budgets.

\begin{table}[ht!]
\centering
\footnotesize
\caption{WN18RR test MRR across edge budgets for the ablation study.}
\label{tab:ablation-wn18rr}
\setlength{\tabcolsep}{3pt}
\begin{tabular*}{\columnwidth}{@{\extracolsep{\fill}}lccccccc}
\toprule
Method & 25 & 50 & 75 & 100 & 300 & 500 & 1000 \\
\midrule
RW & 0.384 & 0.434 & 0.451 & 0.462 & 0.489 & 0.501 & 0.515 \\
RAWExplainer & 0.455 & 0.489 & 0.494 & 0.499 & 0.509 & 0.515 & 0.521 \\
RAWExplainer ($\Phi$) & 0.454 & 0.479 & 0.490 & 0.494 & 0.506 & 0.510 & 0.521 \\
RAWExplainer ($\Phi^{unif}_{eval}$) & 0.462 & 0.483 & 0.492 & 0.496 & 0.507 & 0.511 & 0.522 \\
RAWExplainer (-RW) & 0.420 & 0.436 & 0.443 & 0.446 & 0.439 & 0.433 & 0.435 \\
RAWExplainer (-RelEmb) & 0.453 & 0.481 & 0.491 & 0.497 & 0.504 & 0.511 & 0.521 \\
RAWExplainer (0Topk) & 0.406 & 0.439 & 0.454 & 0.465 & 0.491 & 0.503 & 0.517 \\
RW (2Topk-0.2TP-SumAdj) & 0.457 & 0.483 & 0.492 & 0.498 & 0.504 & 0.511 & 0.521 \\
\bottomrule
\end{tabular*}
\end{table}

\paragraph{(a) Evaluator choice for RAW-Explainer}
Under a fixed scoring protocol, RAW-Explainer (trained with $\Phi^{dist}_{eval}$) generally outperforms variants trained with $\Phi$ and $\Phi^{unif}_{eval}$, except at the 25- and 1000-edge thresholds.

\paragraph{(b) Impact of Random Walk}
Removing the Random Walk (RW) component and $\mathcal{L}_{PPR}$ objective (``RAW-Explainer (-RW)'') significantly reduces performance, confirming the RW block's necessity.

\paragraph{(c) Relation Embeddings}
Removing explicit relation information (``RAW-Explainer (-RelEmb)'') slightly reduces performance, reinforcing the importance of modeling heterogeneity. We attribute the relatively small size of this performance gap to the architecture of the underlying model. NBFNet inherently encodes heterogeneous information into each latent node representation $h_i$. Consequently, even when explicit relation features are removed from the explainer, the representations $h_i$ and $h_j$ of neighboring nodes already implicitly capture the underlying relational context.

\paragraph{(d) RAW-Explainer with 0 Top-$k$ Tails}
RAW-Explainer with 0 top-$k$ tails (``RAW-Explainer (0Topk)'') outperforms the RW baseline (which also lacks them), highlighting our learning component.

\paragraph{(e) Random Walk baseline with the same settings as RAW-Explainer}
Our intent in including the RW baseline is to provide a naive reference point that does not utilize any information from the backbone model, i.e., the top-$k$ tails. However, a more direct comparison is necessary to eliminate any confounding factors. To address this, we conduct an additional experiment that evaluates the RW baseline under the exact same settings as RAW-Explainer, labeled as ``RW (2Topk-0.2TP-SumAdj)''.

As shown in Table~\ref{tab:ablation-wn18rr}, under matched settings, RAW-Explainer consistently matches or outperforms the strengthened RW baseline across most edge configurations, with the sole exception of the 25-edge threshold.

\section{Backbone Sensitivity}
\label{app:backbone-sensitivity}

Our main evaluation focuses on NBFNet as a state-of-the-art path-based model. We conduct an additional set of experiments using RED-GNN to extend our discussion. Table~\ref{tab:redgnn-mrr} reports the results on WN18RR and NELL-995.

\begin{table*}[htbp]
\centering
\footnotesize
\caption{Test MRR across edge budgets on WN18RR and NELL-995 with the RED-GNN backbone.
\textbf{Best} and \underline{second best} results are highlighted within each dataset.}
\label{tab:redgnn-mrr}
\setlength{\tabcolsep}{3pt}
\begin{tabular*}{\textwidth}{@{\extracolsep{\fill}}lccccccc}
\toprule
Method & 25 & 50 & 75 & 100 & 300 & 500 & 1000 \\
\midrule
\multicolumn{8}{c}{\textbf{WN18RR}} \\
\midrule
RW & \textbf{0.391} & \textbf{0.439} & \textbf{0.455} & \underline{0.466} & \textbf{0.495} & \textbf{0.508} & \textbf{0.518} \\
BFS & 0.027 & 0.031 & 0.032 & 0.033 & 0.039 & 0.042 & 0.051 \\
GNNExplainer & 0.227 & 0.334 & 0.367 & 0.382 & \underline{0.397} & 0.401 & 0.403 \\
PGExplainer & 0.234 & 0.312 & 0.345 & 0.359 & 0.396 & 0.408 & 0.423 \\
PaGELink & \underline{0.343} & 0.360 & 0.362 & 0.362 & 0.362 & 0.362 & 0.362 \\
PowerLink & 0.312 & 0.325 & 0.326 & 0.326 & 0.326 & 0.326 & 0.326 \\
RAWExplainer & 0.339 & \underline{0.417} & \underline{0.452} & \textbf{0.467} & \textbf{0.495} & \underline{0.504} & \underline{0.512} \\
\midrule
\multicolumn{8}{c}{\textbf{NELL-995}} \\
\midrule
RW & \underline{0.294} & \textbf{0.345} & \underline{0.367} & \underline{0.378} & \underline{0.402} & \underline{0.416} & \underline{0.432} \\
BFS & 0.021 & 0.023 & 0.040 & 0.041 & 0.080 & 0.083 & 0.089 \\
GNNExplainer & 0.166 & 0.213 & 0.231 & 0.242 & 0.297 & 0.310 & 0.321 \\
PGExplainer & 0.203 & 0.243 & 0.254 & 0.260 & 0.297 & 0.303 & 0.320 \\
PaGELink & 0.227 & 0.229 & 0.230 & 0.230 & 0.230 & 0.230 & 0.230 \\
PowerLink & 0.197 & 0.198 & 0.198 & 0.198 & 0.198 & 0.198 & 0.198 \\
RAWExplainer & \textbf{0.309} & \underline{0.343} & \textbf{0.380} & \textbf{0.393} & \textbf{0.414} & \textbf{0.426} & \textbf{0.442} \\
\bottomrule
\end{tabular*}
\end{table*}

On WN18RR, the results indicate that RW serves as a very strong baseline, and RAW-Explainer approaches its performance at higher edge budgets. We hypothesize that the slightly limited performance of RAW-Explainer with RED-GNN compared to NBFNet is due to differences in the backbone models' inductive biases: NBFNet prefers more direct path connections, whereas RED-GNN emphasizes relational digraphs. The explainer may be overfitting to the original backbone's specific inductive bias, occasionally leading to suboptimal results. Nevertheless, while other learning-based explanation methods experience a significant drop in performance when switching to RED-GNN, RAW-Explainer remains robust and is competitive with the strong RW baseline.

On NELL-995 (Table~\ref{tab:redgnn-mrr}), although RW remains a strong baseline, RAW-Explainer successfully achieves the highest performance across almost all edge budgets, with the sole exception of the 50-edge threshold. These results substantiate our claim that RAW-Explainer is a robust and highly effective approach compared to other learning-based explainers.

\section{Other Explanation Objectives}
\label{app:other-objectives}
We distinguish explanation objectives along two axes, \emph{phenomenon versus model} and \emph{factual versus counter-factual}~\cite{amara2022graphframex}. Phenomenon-focused explanations seek graph patterns that support the ground-truth label. Model-focused explanations seek to account for the original model's behavior. Factual explanations identify a subgraph that supports the target on its own; counter-factual explanations identify edges whose removal weakens support for that target~\cite{amara2022graphframex, yuan2021explainability}.

Our work focuses on \emph{phenomenon-focused factual explanations} to uncover the graph patterns underlying true relationships in knowledge graph completion (see Section \ref{prelim}). We evaluate how well the selected subgraphs support the true entities across varying edge budgets. Nevertheless, other explanation objectives, specifically model-focused factual/counter-factual explanations and phenomenon-focused counter-factual explanations, could provide a complementary perspective to our method.

For model-focused evaluation, we adapt fidelity metrics~\cite{yuan2021explainability} to measure per-query deviations from the original model performance on the full graph. Let $N$ be the number of evaluated queries and $\mathrm{RR}_i(\mathcal{G'})$ the reciprocal rank of the true entity for query $i$ when the original model $\Phi$ uses the graph $\mathcal{G'}$. For each query, $\mathcal{G}$ is the full graph, $\mathcal{G}_{S}$ is its explanatory subgraph, and $\mathcal{G}_{\setminus S}$ is the original graph with the explanatory edges removed. We define
\begin{equation}
    \mathrm{fid}_{-} = \frac{1}{N}\sum_{i=1}^{N} \left|\mathrm{RR}_i(\mathcal{G}) - \mathrm{RR}_i(\mathcal{G}_{S})\right|
\end{equation}
\begin{equation}
    \mathrm{fid}_{+} = \frac{1}{N}\sum_{i=1}^{N} \left|\mathrm{RR}_i(\mathcal{G}) - \mathrm{RR}_i(\mathcal{G}_{\setminus S})\right|.
\end{equation}

Lower $\mathrm{fid}_{-}$ indicates that the explanatory subgraphs better preserve the original model performance in terms of per-query reciprocal ranks, assessing model-focused factual explanations. Higher $\mathrm{fid}_{+}$ indicates a larger deviation from the original model performance after edge removal, assessing model-focused counter-factual explanations.

For phenomenon-focused evaluation, the $\mathrm{MRR}(\mathcal{G}_S)$ used in the main experiments evaluates factual explanations, with higher values indicating stronger support for the true entities within the explanatory subgraphs. We also report $\mathrm{MRR}(\mathcal{G}_{\setminus S})$ to evaluate counter-factual explanations, with lower values indicating weaker support for the true entities after removing the explanatory edges. Both MRR scores average the true entities' reciprocal ranks over all evaluated queries, using the corresponding graph input for each query.

We evaluate these explanation objectives on WN18RR using NBFNet as the original model $\Phi$, following the experimental settings in Section~\ref{ssec:mainres}.

\subsection{Results}

RAW-Explainer achieves the lowest $\mathrm{fid}_{-}$ at all evaluated budgets from 25 to 100 edges (Table~\ref{tab:fidminus-wn18rr}), reaching $0.126$ at 50 edges. Its $\mathrm{fid}_{-}$ increases to $0.181$ at 1000 edges, so a larger explanatory subgraph does not always preserve the full-graph reciprocal ranks more closely. PaGE-Link has the lowest $\mathrm{fid}_{-}$ at 300, 500, and 1000 edges, with a value of $0.139$ at each budget. PaGE-Link’s unchanged scores at larger budgets reflect its path-finding timeout, which can prevent it from using the full edge budget.

The edge-removal evaluations also favor RAW-Explainer at budgets of 25--100 edges. It achieves both the highest $\mathrm{fid}_{+}$ (Table~\ref{tab:fidplus-wn18rr}) and the lowest $\mathrm{MRR}(\mathcal{G}_{\setminus S})$ (Table~\ref{tab:mrrcompl-wn18rr}) at each of these budgets. Removing its selected edges therefore produces the largest average reciprocal-rank deviation and leaves the lowest MRR for the true entities among the compared methods. At 300, 500, and 1000 edges, PG-Explainer leads both measures. RAW-Explainer ranks second at 300 and 500 edges; at 1000 edges, RW narrowly surpasses it on both $\mathrm{fid}_{+}$ ($0.541$ versus $0.540$) and $\mathrm{MRR}(\mathcal{G}_{\setminus S})$ ($0.017$ versus $0.018$). The leading method at larger budgets thus depends on whether the objective is to preserve reciprocal ranks within the selected subgraph or to disrupt them by removing its edges. The main experiments and these evaluations show that RAW-Explainer performs strongly across both the phenomenon/model and factual/counter-factual axes, particularly at small edge budgets.

\begin{table}[ht!]
\centering
\footnotesize
\caption{WN18RR test $\mathrm{fid}_{-}$ across edge budgets. Lower is better.}
\label{tab:fidminus-wn18rr}
\setlength{\tabcolsep}{3pt}
\begin{tabular*}{\columnwidth}{@{\extracolsep{\fill}}lccccccc}
\toprule
Method & 25 & 50 & 75 & 100 & 300 & 500 & 1000 \\
\midrule
RW & 0.180&	0.150&	0.148&	0.148&	0.166&	0.174&	0.185 \\
BFS & 0.505&	0.500&	0.499&	0.498	&0.493	&0.491	&0.484 \\
GNNExplainer & 0.151&	0.148&	0.151&	0.154&	0.158	&0.159&	0.164 \\
PGExplainer & 0.148&	\underline{0.136}&	\underline{0.136}&	\underline{0.136}&	0.158	&0.169	&0.179 \\
PaGELink & \underline{0.139}&	0.139	&0.139&	0.139&	\textbf{0.139}	&\textbf{0.139}	&\textbf{0.139} \\
PowerLink & 0.149	&0.150	& 0.150 &	0.150&	\underline{0.150}&	\underline{0.150}	&\underline{0.150} \\
RAWExplainer & \textbf{0.128}&	\textbf{0.126}&	\textbf{0.130}	&\textbf{0.135}	&0.157	&0.166	&0.181 \\
\bottomrule
\end{tabular*}
\end{table}

\begin{table}[ht!]
\centering
\footnotesize
\caption{WN18RR test $\mathrm{fid}_{+}$ across edge budgets. Higher is better.}
\label{tab:fidplus-wn18rr}
\setlength{\tabcolsep}{3pt}
\begin{tabular*}{\columnwidth}{@{\extracolsep{\fill}}lccccccc}
\toprule
Method & 25 & 50 & 75 & 100 & 300 & 500 & 1000 \\
\midrule
RW & 0.472 & \underline{0.508} & \underline{0.516} & 0.520 & 0.528 & 0.531 & \underline{0.541}  \\
BFS & 0.063 & 0.069 & 0.071 & 0.071 & 0.080 & 0.085 & 0.097 \\
GNNExplainer & 0.424 & 0.450 & 0.458 & 0.467 & 0.483 & 0.489 & 0.496 \\
PGExplainer & \underline{0.484} & 0.507 & \underline{0.516} & \underline{0.523} & \textbf{0.538} & \textbf{0.540} & \textbf{0.547} \\
PaGELink & 0.438 & 0.444 & 0.447 & 0.449 & 0.451 & 0.451 & 0.451 \\
PowerLink & 0.404 & 0.407 & 0.410 & 0.410 & 0.410 & 0.410 & 0.410\\
RAWExplainer & \textbf{0.500} & \textbf{0.522} & \textbf{0.527} & \textbf{0.528} & \underline{0.532} & \underline{0.534} & 0.540 \\
\bottomrule
\end{tabular*}
\end{table}

\begin{table}[ht!]
\centering
\footnotesize
\caption{WN18RR test $\mathrm{MRR}(\mathcal{G}_{\setminus S})$ across edge budgets. Lower is better.}
\label{tab:mrrcompl-wn18rr}
\setlength{\tabcolsep}{3pt}
\begin{tabular*}{\columnwidth}{@{\extracolsep{\fill}}lccccccc}
\toprule
Method & 25 & 50 & 75 & 100 & 300 & 500 & 1000 \\
\midrule
RW & 0.094 & \underline{0.054} & \underline{0.045} & \underline{0.040} & 0.031 & 0.027 & \underline{0.017} \\
BFS & 0.490 & 0.485 & 0.482 & 0.482 & 0.475 & 0.471 & 0.463 \\
GNNExplainer & 0.157 & 0.132 & 0.119 & 0.114 & 0.090 & 0.083 & 0.075 \\
PGExplainer & \underline{0.088} & 0.060 & 0.048 & 0.042 & \textbf{0.025} & \textbf{0.018} & \textbf{0.011} \\
PaGELink & 0.133 & 0.126 & 0.124 & 0.121 & 0.120 & 0.120 & 0.120  \\
PowerLink & 0.166 & 0.163 & 0.162 & 0.162 & 0.161 & 0.162 & 0.161\\
RAWExplainer & \textbf{0.071} & \textbf{0.044} & \textbf{0.037} & \textbf{0.034} & \underline{0.027} & \underline{0.025} & 0.018  \\
\bottomrule
\end{tabular*}
\end{table}

\section{Case Study}
\label{case-study}
\begin{figure}[!t]
  \centering
  \includegraphics[width=\columnwidth]{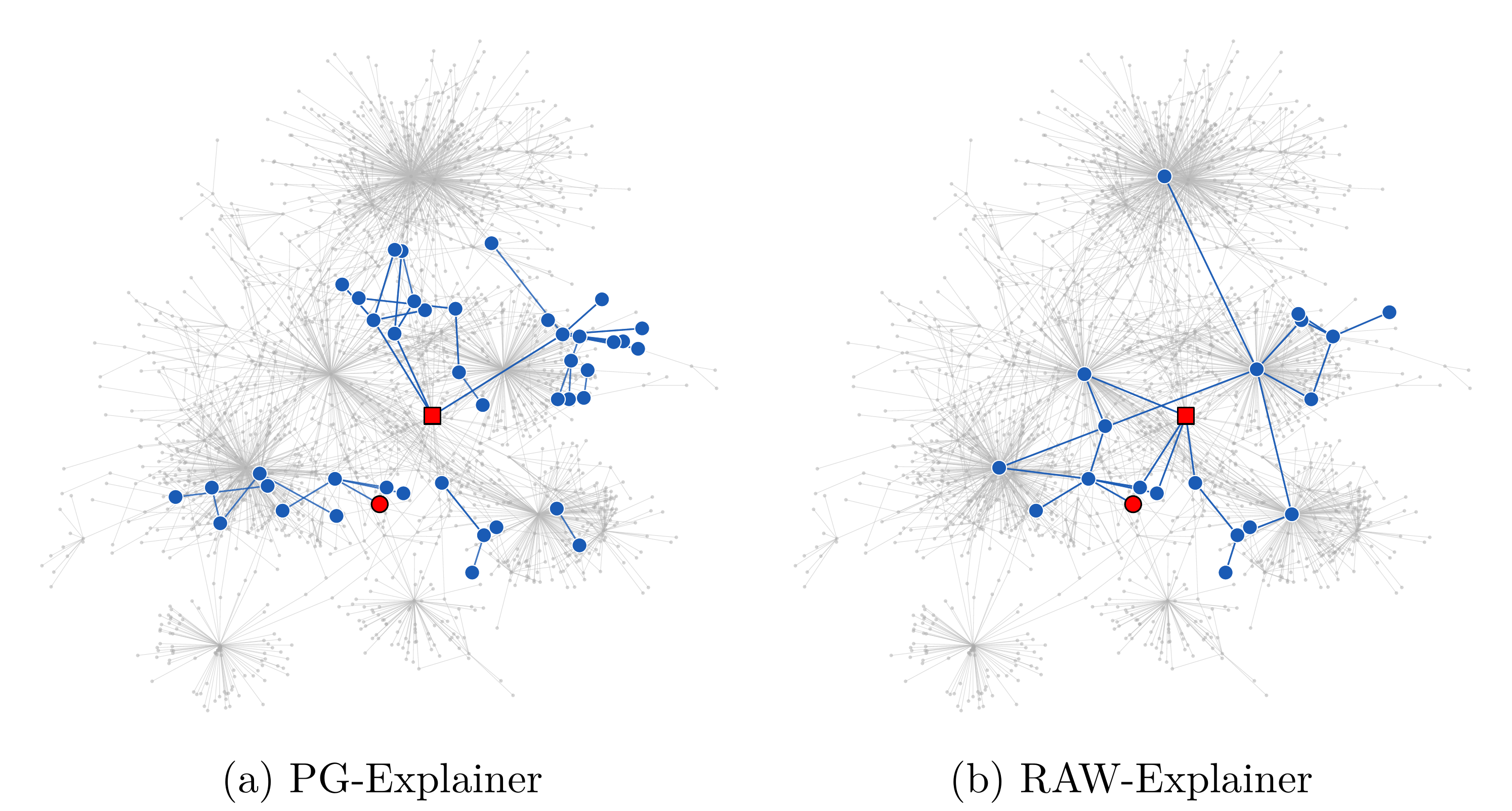}
    \caption{Comparison of explanation subgraphs for a sample WN18RR query with a 50-edge budget (shown in blue; multi-relational and inverse edges are collapsed; the red square and red circle denote the query source and target entities, respectively). (a) PG-Explainer yields a fragmented explanation with 13 connected components. (b) RAW-Explainer produces a denser subgraph with a single connected component, concentrated around the query head and tail entities.}
  \label{fig:explanation_comparison}
  \vspace{-7.5pt}
\end{figure}

To illustrate the qualitative differences between RAW-Explainer and a strong baseline, such as PG-Explainer, we analyze the example WN18RR triple \textcolor{charcoal}{(\textsf{trademark.n.02}, \textsf{member\_of\_domain\_usage}, \textsf{rolaids.n.01})} with a 50-edge budget. \autoref{fig:explanation_comparison} visualizes the explanation subgraphs generated by both methods, using the underlying knowledge graph structure extracted from the WN18RR dataset (limited to 4 hops from the query entity for readability). A key distinction between the two approaches lies in the connectivity of the explanation subgraphs. PG-Explainer produces 13 disconnected components, whereas RAW-Explainer yields a single connected component within budget.

The explanatory subgraph of PG-Explainer indicates that the model identifies numerous isolated nodes and edges that collectively contribute to the model's predictive performance. However, this fragmentation makes it difficult to reason about the contribution of each edge to the prediction. In contrast, RAW-Explainer produces a more structurally compact explanation, which can support human interpretability through path reasoning and may reduce the cognitive effort needed to understand the semantic relationship between the explanation and the prediction.

\makeatletter
\newlength{\savedfptop}
\setlength{\savedfptop}{\@fptop}

\setlength{\@fptop}{0pt}
\makeatother

\section{Limitations}
\label{app:limitations}

Our evaluation focuses primarily on NBFNet, with additional experiments using RED-GNN on WN18RR and NELL-995 (\autoref{app:backbone-sensitivity}). These experiments extend the evaluation to a second backbone, but generalization to a wider range of GNN architectures remains to be established.

\clearpage

\makeatletter
\setlength{\@fptop}{\savedfptop}
\makeatother

\end{document}